\documentclass[letterpaper, 10 pt, conference]{ieeeconf}  

\IEEEoverridecommandlockouts                               

\usepackage{graphics} 
\usepackage{epsfig} 
\usepackage{amsmath} 
\usepackage{amssymb}  
\usepackage{comment}
\usepackage{xcolor}
\usepackage{booktabs}
\usepackage{siunitx}
\usepackage{algorithm} 
\usepackage{algpseudocode}
\usepackage{hyperref}
\usepackage{cite}
\DeclareMathAlphabet{\mathcalligra}{T1}{calligra}{m}{n}

\title{\LARGE \bf
 Robot Learning with Vision-Language-Action Models for Agricultural Manipulation
}

\author{Hamid Hafezi$^{1}$, Javad Mohammadpour Velni$^{2}$, Mohammadreza Davoodi$^{1}$
\thanks{*This material is based upon work supported by the National Science Foundation (NSF) under award No. 2330216.}
\thanks{$^{1}$Department of Electrical and Computer Engineering, The University of Memphis, Memphis, TN, USA
        {\tt\small hhafezi@memphis.edu, mdavoodi@memphis.edu}}
\thanks{$^{2}$Department of Mechanical Engineering, Clemson University, Clemson, SC 29634 USA
        {\tt\small javadm@clemson.edu}}}

\begin{document}

\maketitle
\thispagestyle{empty}
\pagestyle{empty}

\begin{abstract}
Robotic manipulation in agricultural environments remains challenging due to variations in plant geometry, occlusion, and camera viewpoints. Pretrained vision-language-action (VLA) and generalist robot policies offer a promising approach to addressing these challenges by learning manipulation behaviors from multimodal observations and language instructions. However, adapting these policies to agricultural tasks requires domain-specific demonstrations, which are costly and time consuming to collect on physical robots.
In this work, we develop a simulation-to-real robot learning framework for agricultural reaching and grasping tasks that integrates task-specific demonstration collection and pretrained policy adaptation. A MuJoCo-based agricultural environment is developed with diverse tomato plant structures to support demonstration collection and controlled evaluation. Expert demonstrations data are collected through joystick teleoperation using a low-level controller that executes the learned policy actions. Two pretrained generalist robot policies, namely, Octo and $\pi_{0.5}$, are adapted using simulated demonstrations, followed by further fine-tuning on a physical Franka Research 3 manipulator using real robot demonstrations.
The adapted policies are evaluated under in-distribution, viewpoint/spatial-shift, and out-of-distribution conditions in simulation, followed by physical robot experiments. The influence of proprioceptive information on policy performance is also investigated. Results show that $\pi_{0.5}$ achieves higher task-performance scores than Octo and exhibits less degradation under distribution shifts. 
These findings underline the potential of integrating pretrained generalist policies with task-specific demonstration learning to enhance robotic manipulation in agricultural applications.
\end{abstract}

\begin{keywords}
Robot Learning; Vision Language Action; Agricultural Tasks; Learning from Demonstration.
\end{keywords}

\section{INTRODUCTION}

Many agricultural operations remain labor intensive and physically demanding, making them a compelling application domain for robot-assisted automation. Robotic manipulators are increasingly being explored for tasks such as harvesting, planting, monitoring, weeding, and spraying in controlled environments (such as greenhouses) and open fields \cite{jin2024robotic,jo2025review}.
However, agricultural environments are substantially less structured than typical industrial settings. Crops exhibit significant variations in geometry, orientation, and height, while leaves, branches, and other plant structures can occlude targets and constrain manipulator access \cite{rong2025decoupled,zhao2026improved}. These characteristics make target localization, motion planning, and reliable robotic manipulation complicated.

\begin{figure}[tp]
  \centerline{\includegraphics[width=0.5\textwidth]{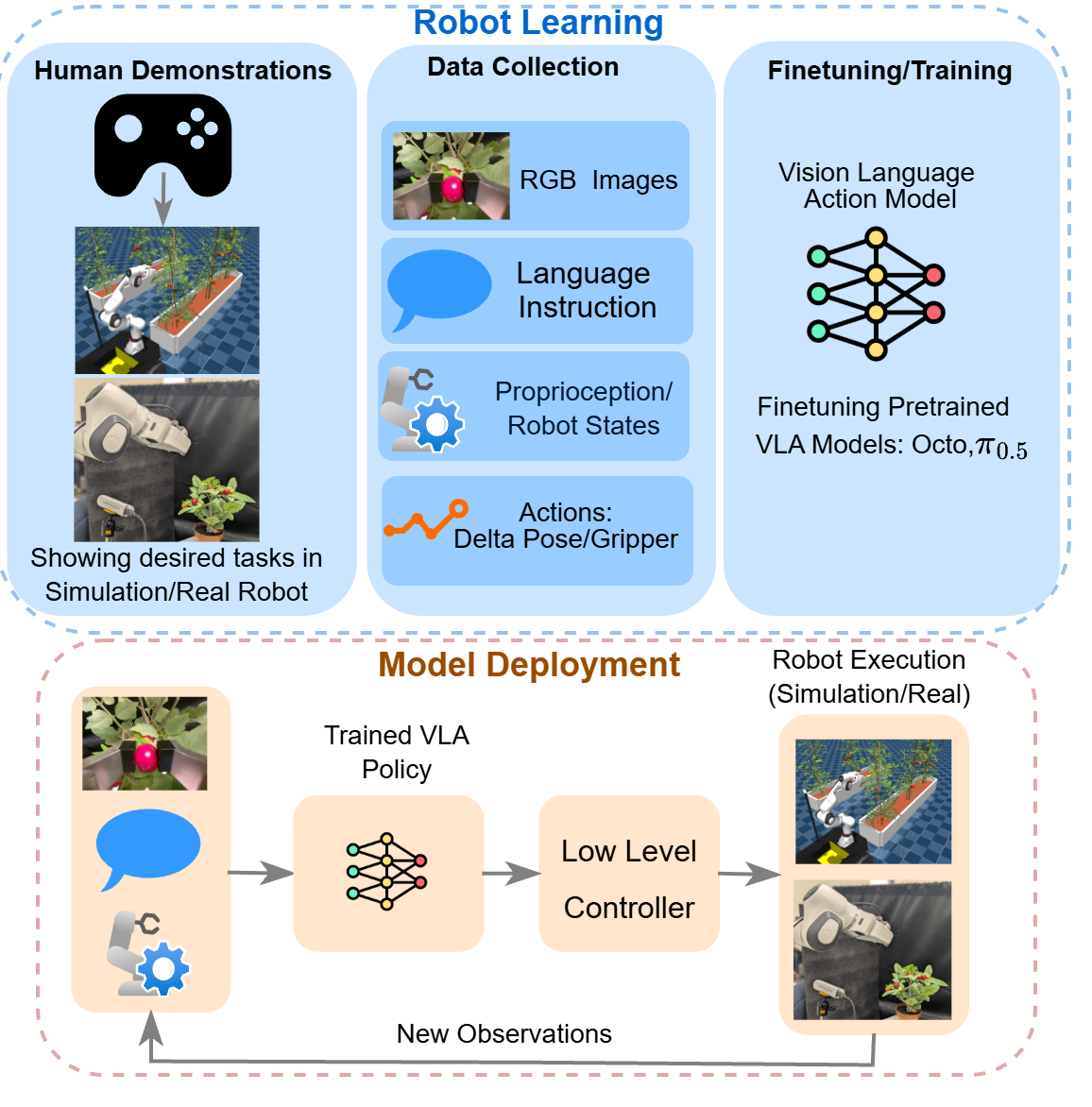}}
  \caption{General structure of end-to-end robot learning: demonstrating
the desired task to the robot, fine-tuning or training a policy, and deploying the policy for task execution.}
    \vspace{-1.5em}
  \label{fig:general_structure}
\end{figure}

Conventional agricultural manipulation systems commonly employ
a modular architecture in which perception, planning, and control
are implemented as separate components \cite{jin2024robotic}.
On the perception side, the literature commonly relies on vision or multi-sensor processing to detect crops and estimate 3D target locations, often requiring careful camera placement, calibration, and coordinate-frame alignment to achieve reliable target estimates under occlusion and clutter \cite{liu2026deep,gursoy2023towards}. For example, in \cite{miao2023efficient}, a tomato harvesting system combines YOLOv5 with image processing and depth information to recognize tomato clusters and localize stems, transforming image-based estimates into spatial coordinates for robot motion. Similarly, in \cite{gursoy2023towards}, a dual-arm fruit harvesting system combines YOLOv5 and object tracking with RGB-D measurements to estimate fruit positions, while image processing identifies the tree trunk.
In these systems, learning-based perception serves as a front-end component that provides geometric targets and information about the surrounding scene for downstream motion generation.

Using these target estimates and a representation of the workspace, the planning module then generates feasible motions that account for the robot's kinematics, reachability, and surrounding obstacles.
In agricultural applications, \cite{rong2025decoupled} combines manipulator placement optimization with an improved RRT* planner to generate collision-free trajectories for tomato harvesting.
Other fruit picking systems modify the RRT* sampling strategy to improve path generation in cluttered environments containing branches and leaves \cite{zhao2026improved}. During motion execution, feedback control can improve positioning accuracy and interaction safety. In \cite{shi2023lightweight}, visual feedback is used to refine the end-effector position, while estimated contact forces are monitored to prevent unsafe interaction with workers or plants. In \cite{wang2023biologically}, localized fruit positions
are used by a neural control framework to generate coordinated reaching and cutting motions while accounting for robot constraints. These studies illustrate how conventional agricultural manipulation systems integrate perception, planning, and control through a modular architecture.

Although such modular approaches benefit from mature robotics tools and explicit handling of constraints, their performance depends on the accuracy and completeness of the information exchanged between components. 
Errors in target localization or incomplete representations of surrounding plant structures can affect motion generation and execution. Reported challenges include occluded targets, unmodeled branches, and fruit displacement during interaction \cite{gursoy2023towards}.
Changes in lighting and crop arrangement may also require adjustments to perception and target-access strategies \cite{miao2023efficient}.
Moreover, many agricultural robotic systems are developed for specific crops, environments, and tasks, and adapting them to new agricultural settings may require additional development and validation \cite{jin2024robotic,jo2025review}.
These limitations motivate the investigation of learning-based approaches that map observations more directly to robot actions while retaining feedback and low-level control for execution.

Robot learning provides a framework for acquiring manipulation behaviors from data and interaction rather than relying solely on manually designed models and controllers \cite{billard2016learning,calinon2019learning}.
Two major paradigms are reinforcement learning (RL), which learns behaviors through interaction and reward feedback \cite{kober2013reinforcement}, and learning from
demonstration (LfD), which learns from expert demonstrations \cite{billard2016learning}. Recent advances in machine learning and the growing availability of large-scale robot datasets have significantly expanded both paradigms \cite{o2024open}. In particular, pre-collected datasets have enabled offline RL, while large and diverse demonstration datasets have supported the development of generalist policies that can transfer across tasks, environments, and robot embodiments. While offline RL typically relies on reward information associated with collected trajectories, specifying appropriate reward functions for complex manipulation tasks can be challenging. In contrast, LfD enables policies to be learned directly from expert demonstrations without requiring explicit reward labels \cite{shafiullah2022behavior}.

Previous research has explored a wide range of LfD approaches for acquiring robot manipulation skills. One class of methods focuses on representing and generalizing demonstrations at the trajectory level (e.g., joint-space or end-effector motion) using structured movement representations such as probabilistic and neural network-based movement primitives \cite{calinon2007learning, saveriano2021dynamic, seker2019conditional, theofanidis2024safe, ravichandar2020recent, hafezi2026safe}. A second class instead learns policies that directly map robot observations, such as RGB/depth images and proprioceptive states, to robot actions, including joint-space, end-effector, or gripper commands. Representative examples include implicit policies formulated as energy-based models \cite{florence2022implicit}, diffusion-based visuomotor policies \cite{chi2025diffusion}, and Transformer-based behavior cloning approaches such as the Behavior Transformer \cite{shafiullah2022behavior}. Although these methods have demonstrated strong performance, many are developed for specific tasks and trained on datasets with limited diversity. More recently, large-scale multi-task robot datasets such as Open X-Embodiment \cite{o2024open} have accelerated interest in generalist robot policies, which use large Transformer backbones to scale across tasks and embodiments. Some of these policies, such as RT-1 \cite{brohan2022rt} and Octo \cite{team2024octo}, are trained primarily on robot data with lightweight vision and language encoders. Others, referred to as vision-language-action (VLA) models, build on vision-language models pretrained on web-scale data and adapt them to predict robot actions; examples include RT-2 \cite{brohan2023rt}, OpenVLA \cite{kim2024openvla}, $\pi_{0}$ \cite{black2024pi_0}, and $\pi_{0.5}$ \cite{intelligence2025pi0}. Both families have demonstrated broad transfer and, in some settings, zero-shot or few-shot generalization to novel tasks.

Recent studies have also investigated learning-based and vision-language models (VLMs) for agricultural robotics, including task-specific imitation learning for outdoor pepper harvesting \cite{kim2025autonomous}, VLM-guided task planning for crop monitoring through navigation and active-vision tools \cite{cuaran2026visual}, and scene-graph-grounded vision-language reasoning for greenhouse manipulation through predefined basic actions \cite{park2026scene}.

Despite these advances, adapting pretrained generalist policies to agricultural manipulation remains limited and challenging due to differences between pretraining and agricultural environments, while collecting task-specific demonstrations across diverse plant configurations on physical robots is costly and time consuming. Moreover, understanding how these adapted policies perform under variations in plant structure, camera viewpoint, and robot configuration remains an important research question. To address these challenges, we develop a simulation-to-real robot learning framework that integrates demonstration collection, pretrained policy adaptation, and low-level manipulator control. The framework includes a MuJoCo-based agricultural simulation environment for scalable demonstration collection and controlled evaluation, together with a joystick-based teleoperation system and a damped, null-space-based inverse kinematics controller for demonstration collection and learned-policy execution. Two pretrained generalist robot policies, Octo and $\pi_{0.5}$, are adapted to a tomato reaching-and-grasping task through fine-tuning on simulated demonstrations, followed by further fine-tuning using demonstrations collected on a physical Franka Research 3 (FR3) robot. We further evaluate the adapted policies under in-distribution, viewpoint/spatial-shift, and out-of-distribution conditions in simulation, followed by physical robot experiments, while investigating the influence of proprioceptive information on policy performance. The overall learning and deployment workflow is illustrated in Fig.~\ref{fig:general_structure}.

The main contributions of this work are as follows:
\begin{itemize}
    \item A MuJoCo-based agricultural manipulation environment with diverse tomato plant structures, integrated with a joystick-based teleoperation and low-level inverse-kinematics control framework for demonstration collection and
    learned policy execution.

    \item A simulation-to-real adaptation of two pretrained generalist robot policies, Octo and $\pi_{0.5}$, for agricultural reaching and grasping, using simulated demonstrations followed by further fine-tuning with demonstrations collected on a physical Franka Research 3 (FR3) robot.

    \item A comparative evaluation of the adapted policies under in-distribution, viewpoint/spatial-shift, and out-of-distribution conditions in simulation, together with real robot experiments investigating their performance and the influence of proprioceptive information.
\end{itemize}


\section{Task Formulation and Preliminaries }

\begin{figure*}[tp]
  \centerline{\includegraphics[ height=0.25\textheight]{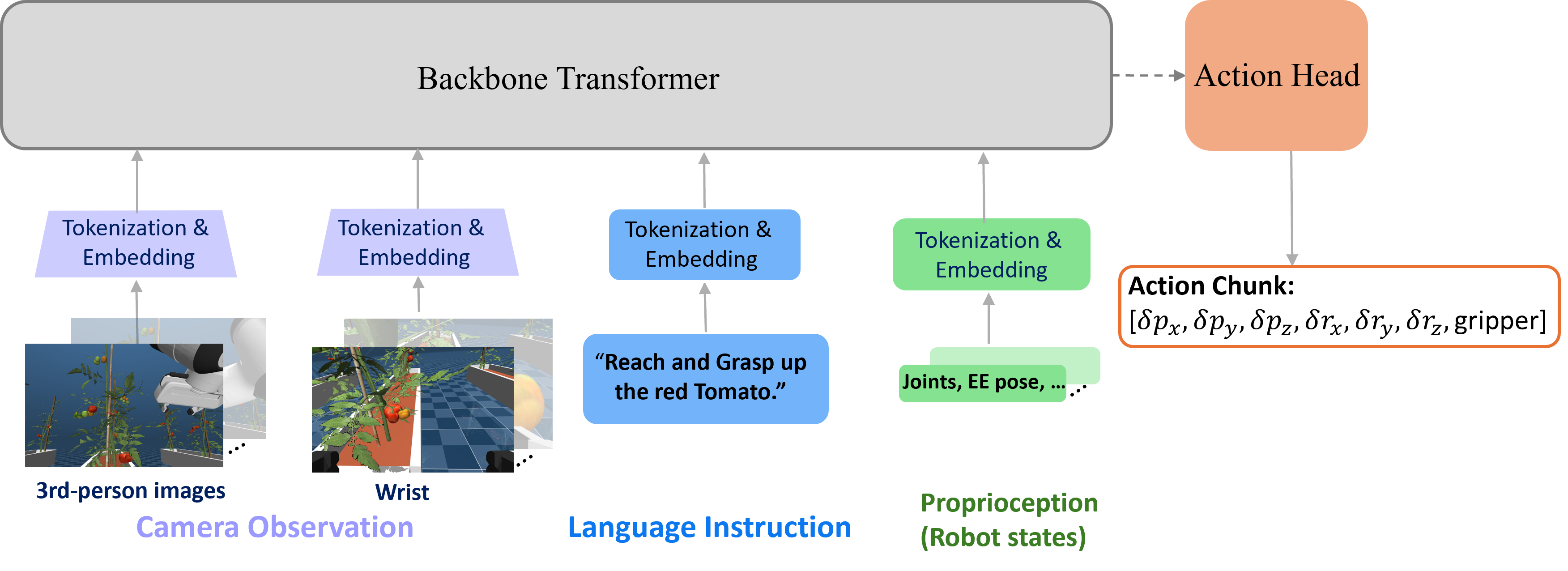}}
  \caption{General structure of Vision Language Action models}
  \label{fig:vla_structure}
\end{figure*}

\subsection{Agricultural Manipulation Task Formulation}
\label{subsec:task_formulation}

We consider a robot-assisted reaching-and-grasping task in which
a manipulator must approach and grasp a target ripe tomato
under variations in plant geometry, fruit location, and
visual occlusion. At each time step $t$, the policy receives
visual observations from a primary (third-person) camera and a wrist-mounted camera, together with a natural language task instruction and, depending on the experimental configuration, proprioceptive information. Based on these
inputs, the policy predicts a sequence of end-effector
motion and gripper commands to accomplish the task.
This task is representative of agricultural manipulation because
it requires target localization, precise end-effector positioning,
and interaction in cluttered and partially occluded environments.
These capabilities are also relevant to other agricultural
operations, such as pruning, monitoring, and spraying.
We investigate the adaptation of two pretrained generalist
robot policies, Octo and $\pi_{0.5}$, to this task. The following subsections
introduce the general VLA formulation and the
architectures of the two models.  Section \ref{sec: env} presents the MuJoCo simulation environment
and experimental setup, including demonstration collection
and evaluation scenarios, while Section \ref{section:finetunining_evaluation} describes
the policy fine-tuning procedures, experimental evaluation,
and results.
\color{black}

\subsection{Why Vision-Language-Action (VLA) Models?}

VLA models enable robots to learn policies that directly map multimodal observations and natural language instructions to executable robotic actions. They are commonly trained through behavior cloning using demonstrations collected from human operators, scripted controllers, or previously trained policies. In contrast to conventional modular pipelines, where perception, planning, and control are designed separately, VLA policies learn an observation-to-action mapping from
demonstration data while relying on a low-level controller for execution. This approach is particularly relevant to
agricultural manipulation, where variations in plant structure and fruit placement can complicate explicit
perception and motion planning.

To frame the problem mathematically, let $\mathcal{D}=\{\zeta_i\}_{i=1}^{N_{\mathcal{D}}}$ denote a dataset of expert demonstrations, where $\zeta_i=\{(o^i_t,l^i,a^i_t)\}_{t=0}^{T_i-1}$ represents the $i$-th trajectory of length $T_i$. A VLA policy generates the robot action $a_t\in\mathbb{R}^{d_a}$ conditioned on an observation history
$o_{t-H_o+1:t}$ of length $H_o$ and a natural language instruction $l$, expressed as $a_t\sim\pi(\cdot|o_{t-H_o+1:t},l)$. 
The robot observations $o_t$ may include camera images $I_t$ and proprioceptive state $s_t$, which contains low-dimensional information such as joint positions, joint velocities, and end-effector pose.
The language instruction $l$ is a text-based input that specifies the desired task. The robot action $a_t\in\mathbb{R}^{d_a}$ denotes the continuous command applied through the robot control interface, which may represent joint-space commands, Cartesian end-effector commands, or gripper actions, depending on the application. However,  in many robot learning datasets, actions are represented as incremental changes in joint configurations or end-effector poses. Recent robot policies also employ action chunking \cite{zhao2023learning}, in which a sequence of $H_a$
consecutive actions, $a_{t:t+H_a-1}$, is predicted instead of a single action $a_t$. Predicting action sequences can help reduce compounding errors during sequential execution \cite{zhao2023learning, capuano2025robot}. Recent VLA models employ diffusion or flow-matching methods to jointly generate continuous action chunks rather than predicting actions sequentially.

Despite architectural differences, many VLA models share a common three-component structure, as illustrated in Fig. \ref{fig:vla_structure}: (i) modality-specific tokenizers/encoders that map the language instruction and each observation into a shared embedding space; (ii) a Transformer backbone that fuses multimodal tokens; and (iii) an action head that generates an action chunk
$a_{t:t+H_a-1}$. Beyond this shared structure, models differ in several
key aspects. Some models train their Transformer backbones primarily from scratch on robot demonstration data, whereas others initialize them from VLMs pretrained on web-scale image-text data and thus inherit broader semantic priors. 
Models also vary in scale, from relatively compact policies to architectures with billions of parameters, with larger models generally requiring more computational resources. They further differ in their tokenization strategies and action-generation mechanisms. In this work, we study two pretrained generalist robot policies, namely, Octo and $\pi_{0.5}$, which differ in their architectures and pretraining strategies.




\subsection{Octo Model Architecture} \label{subsec:octo_arch}

Octo is a Transformer-based generalist robot policy. It is available in two
sizes, Octo-Small (~27M parameters) and Octo-Base (~93M parameters), of which we use Octo-Base for this study. 
Unlike models initialized from pretrained VLM backbones, Octo's Transformer backbone is trained from scratch on robot demonstration data.
The model is pretrained in a single stage on a mixture
of 25 datasets from the Open X-Embodiment collection,
consisting of robot demonstrations with visual observations
and action commands.
Octo processes task and observation tokens through a
Transformer backbone and uses learned readout tokens
to obtain contextual representations for action prediction.
A diffusion action head then generates continuous action
chunks from these representations.


\paragraph{Tokenization}
The language instruction $l$ is tokenized and encoded
using a frozen, pretrained T5-base language encoder
\cite{raffel2020exploring}, as $e_l = \mathrm{T5Enc}\big(\mathrm{Tok}(l)\big)$.
Each camera image $I_t^C$, with
$C \in \{\text{primary},\text{wrist}\}$, is processed
by a shallow convolutional encoder $\phi_c$, and the
resulting features are flattened into a sequence of
visual tokens, $e^{C}_{t} = \mathrm{Flatten}\big(\phi_c(I^C_t)\big)$.
The low-dimensional proprioceptive state $s_t$ is mapped
to tokens using a dedicated tokenizer $\psi$, $e^{s}_{t} = \psi(s_t)$.
Each token group is then linearly projected into the
shared Transformer dimension $d$ and augmented with
group-specific learnable positional embeddings.
The resulting language tokens are given by $z_l = W_l e_l + p_l$, while the observation tokens
at time step $t$ are represented by $z_t^o=W_o e_t^o+p_t^o$ for observations, where
$o\in\{\mathrm{primary},\mathrm{wrist},s\}$, $W_l$ and $W_o$ denote the projection matrices, and $p_l$ and $p_{o_t}$ denote the corresponding learnable positional embeddings. 

\paragraph{Backbone and block-wise attention}
The Transformer backbone processes a sequence consisting of language tokens followed by observation tokens from a window of $H_o$ time steps, together with a learned readout token $z_{R_t}$ at each time step:
\begin{equation}
\boldsymbol{z} =
\Big[\,z_l,\;
\big\{z^{\text{primary}}_k,\;z^{\text{wrist}}_k,\;
z^{\text{proprio}}_k,\;z_{R_k}\big\}_{k=t-H_o+1}^{t}\,\Big].
\end{equation}
A block-wise causal attention pattern is applied to this
sequence. Observation tokens attend to the language tokens and to observations at the same or earlier time steps, but not to future observations. Each readout token attends to the language tokens and observations at its corresponding time step and earlier time steps, while observation tokens do not attend to readout tokens.
The output embedding $c_t$ of the action readout token at the final Transformer layer summarizes the task
instruction and observations up to time step $t$ and serves as the conditioning representation for
the diffusion action head.

\paragraph{Action Generation and Learning Objective}
The action head is a conditional denoising diffusion model
that generates action chunks conditioned on the readout
representation $c_t$. It employs a denoising
network $f_\theta$ to iteratively transform Gaussian noise
into continuous robot actions.
More specifically, the model generates an action chunk
$a_{t:t+H_a-1}\in\mathbb{R}^{H_a\times d_a}$, where $H_a$
is the action horizon and $d_a$ is the action dimension. 
During training, a demonstrated action chunk is flattened
into a vector $x_t^0\in\mathbb{R}^{H_ad_a}$. A diffusion
step $n$ is sampled uniformly from $\{1,\ldots,N\}$,
where $N$ is the number of diffusion steps. Gaussian noise
is then added to the demonstrated action chunk according
to a fixed variance schedule $\beta_1,\ldots,\beta_N\in(0,1)$.
Defining $\alpha_n=1-\beta_n$ and
$\bar{\alpha}_n=\prod_{j=1}^{n}\alpha_j$, the noisy action
is given by
\begin{equation}
x_t^n
=
\sqrt{\bar{\alpha}_n}\,x_t^0
+
\sqrt{1-\bar{\alpha}_n}\,\epsilon,
\qquad
\epsilon\sim\mathcal{N}(0,I).
\end{equation}

The diffusion head takes the noisy action $x_t^n$, diffusion
step $n$, and readout representation $c_t$ as inputs and
predicts the added noise
$\hat{\epsilon}=f_\theta(c_t,x_t^n,n)$.
The model is trained by minimizing a mean-squared
error between the predicted and actual noise,
\begin{equation}
\mathcal{L}_{\mathrm{diff}}
=
\mathbb{E}_{\mathcal{D},n,\epsilon}
\Big[
\big\|\,\hat{\epsilon}-\epsilon\,\big\|_2^2
\Big],
\label{eq:octo_diff_loss}
\end{equation}
During training, the mask excludes padded time steps and
action dimensions that are invalid for the current robot
embodiment.
At inference, action generation starts from Gaussian
noise $x_t^N\sim\mathcal{N}(0,I)$, which is iteratively
denoised using the learned reverse process for
$n=N,\ldots,1$,
\begin{equation}
x_t^{n-1}
=
\frac{1}{\sqrt{\alpha_n}}
\left(
x_t^n - \frac{\beta_n}{\sqrt{1-\bar{\alpha}_n}}\,f_\theta(c_t,x_t^n,n)
\right)
+
\sigma_n\,\xi, 
\label{eq:octo_reverse}
\end{equation}
where $\sigma_s$ controls the noise injected at each reverse step and $\xi\sim\mathcal{N}(0,I)$.
The denoised vector $x_t^0$ is then reshaped into the
action chunk $a_{t:t+H_a-1}$.

\subsection{$\pi_{0.5}$ Model Architecture}
\label{subsec:pi05_arch}

$\pi_{0.5}$ \cite{intelligence2025pi0} is a larger VLA than Octo and, unlike Octo, is built on a
pretrained VLM backbone. It uses a 3B-parameter PaliGemma VLM \cite{beyer2024paligemma}, augmented with a separate
300M-parameter action expert.  
The model employs a co-training strategy using a heterogeneous mixture of robot demonstrations from multiple embodiments, high-level semantic subtask annotations, and multimodal web data, including image captioning, visual question answering, and object localization, to improve its ability to generalize to environments not encountered during training.


The original $\pi_{0.5}$ training procedure consists of
two stages. During pretraining, the model is trained to
predict discrete tokens autoregressively. Depending on the
training example, these tokens represent semantic outputs,
such as captions, answers to questions, object locations,
and subtask descriptions, or discrete action chunks sequentially.
For post-training, the autoregressive token-prediction
objective is retained, while a separate action expert is
introduced and trained to generate continuous action chunks.
Thus, during post-training, the model learns to predict
action chunks in both discrete and continuous forms.

Additionally, $\pi_{0.5}$ supports hierarchical robot tasks
by first predicting a natural-language subtask description
$\hat{l}$ from the current observation $o_t$ and overall
task instruction $l$, and then generating a low-level
action chunk conditioned on the predicted subtask.
In this work, however, we use the continuous low-level
policy directly, conditioned on the agricultural task
instruction.


\paragraph{Tokenization}

Each camera image $I_t^C$, with $C\in\{\mathrm{primary},\mathrm{wrist}\}$, is processed
by PaliGemma\'s SigLIP vision encoder. The resulting visual features are linearly projected into the
language-model embedding space to produce visual tokens $z_t^c$. The language instruction $l$ is tokenized using
the PaliGemma tokenizer, and the resulting token IDs are mapped to embeddings $z^l$ through the language embedding
layer.
The normalized proprioceptive state $s_t$ is discretized into numerical bins and incorporated into the text prompt, where it is processed by the language tokenizer and embedding layer to obtain proprioceptive tokens $z_t^{\mathrm{proprio}}$. This differs from Octo, which uses a dedicated observation tokenizer and projection for proprioceptive information.


During pretraining, $\pi_{0.5}$ uses Frequency-space Action Sequence Tokenization (FAST), which applies a discrete cosine transform, quantization, and byte-pair
encoding to compress continuous action chunks into short
sequences of discrete tokens $z^{\mathrm{FAST}}_{1:M_F}$
for autoregressive prediction, where $M_F$ denotes the
number of FAST tokens. These tokens are mapped to
embeddings through the language embedding layer.
During post-training, the model additionally introduces
continuous action tokens $z^{a,\tau}_{t:t+H_a-1}$, which
are processed by the separate action expert.

Unlike Octo, which adds learned positional
embeddings to each token group, $\pi_{0.5}$ relies on the positional
embeddings within SigLIP for image patches and on the rotary position
embeddings of the Gemma backbone for the combined sequence.

\paragraph{Backbone and attention}
The encoded image, language, and proprioceptive inputs, together with the
discrete action and continuous action tokens, are arranged into a single multimodal
sequence
\begin{equation}
\boldsymbol{z} =
\big[\,\underbrace{z^{\text{primary}}_t,\;z^{\text{wrist}}_t,\;z^{l},\;z^{\text{proprio}}_t,\;
z^{\text{FAST}}_{1:M_F}}_{\text{VLM weights}},\underbrace{\;z^{a,\tau}_{t:t+H_a-1}}_{\text{action expert}}\,\big],
\end{equation}
which is processed jointly by the backbone transformer. Note that, unlike Octo,
the sequence contains observation tokens from the current timestep $t$ only, without a window of past observations.
The token type determines which set of transformer weights processes them. The image, language, proprioceptive, and FAST tokens are handled by the VLM weights, whereas the
continuous action tokens are handled by the separate action-expert weights. Although the two components use different parameter sets, their representations interact through attention.

The attention mechanism follows a structured masking
pattern. The image, language, and proprioceptive tokens
form the prefix of the sequence, where all tokens can
attend to one another regardless of their positions,
known as bidirectional attention. The FAST tokens attend
to this prefix and to preceding FAST tokens through
causal attention. Continuous action tokens attend to
the entire prefix and to one another, but not to the
FAST tokens, preventing the action expert from accessing
the FAST action tokens. Conversely, the prefix
and FAST tokens cannot attend to the continuous action
tokens, allowing information to flow from the VLM
backbone to the action expert.

\paragraph{Action Generation and Learning Objective}

The $\pi_{0.5}$ architecture supports both discrete token
prediction and continuous action generation. The VLM
component $f_\theta^l$ predicts semantic text and
FAST-encoded action tokens autoregressively, while the
action expert $f_\theta^a$ generates continuous action
chunks.



The action expert employs flow matching~\cite{lipman2022flow},
a generative modeling framework closely related to diffusion.
Unlike Octo's diffusion action head, which learns to reverse
a stochastic noising process, flow matching learns a
time-dependent velocity field that transforms Gaussian noise
into continuous actions along a predefined path. This
approach can generate actions with fewer sampling steps than
conventional diffusion methods. Specifically, given a
demonstrated action chunk $a_{t:t+H_a-1}$ and a Gaussian
noise sample $\omega\sim\mathcal{N}(0,I)$, an intermediate
action representation is constructed as
\begin{equation}
a^{\tau,\omega}_{t:t+H_a-1}
= (1-\tau)\,a_{t:t+H_a-1} + \tau\,\omega,
\qquad \tau\in[0,1],
\end{equation}
where $\tau$ denotes the flow-matching time variable,
with $\tau=1$ corresponding to Gaussian noise and $\tau=0$
to the demonstrated action. The velocity along this path
is $a_{t:t+H_a-1}-\omega$, which the action expert
$f_\theta^a$ is trained to predict.

 The original $\pi_{0.5}$ post-training objective combines autoregressive token prediction with flow-matching regression:
\begin{equation}
\begin{split}
\mathbb{E}_{\mathcal{D},\tau,\omega}\bigg[
&\mathrm{H}\big(y_{1:M},f^{l}_\theta(o_t,l)\big) + \\
&\alpha\big\|\, f^{a}_\theta(a^{\tau,\omega}_{t:t+H_a-1},o_t,l)- (a_{t:t+H_a}-\omega)\,\big\|^2
\bigg],
\end{split}
\label{eq:pi05_loss}
\end{equation}

where $\mathrm{H}$ denotes the cross-entropy loss over the target semantic text or FAST action tokens $y_{1:M}$,
the second term represents the flow-matching regression
loss, and $\alpha$ balances the two objectives. Setting
$\alpha=0$ yields an autoregressive objective, as
used during pretraining, whereas $\alpha>0$ corresponds
to post-training, where both objectives are active.


During inference, the original $\pi_{0.5}$ framework can
first predict a semantic subtask $\hat{l}$ autoregressively
and then generate a continuous action chunk by integrating
the learned velocity field, starting from Gaussian noise.
Using Euler integration, the update is given by
\begin{equation}
a_{t:t+H_a-1}^{\tau-\delta}
= a_{t:t+H_a-1}^{\tau}
- \delta\,f_\theta^a\left(
a_{t:t+H_a-1}^{\tau},\tau,o_t,\hat{l}
\right),
\label{eq:pi05_integration}
\end{equation}
where $\delta>0$ denotes the integration step size.
Starting from $a_{t:t+H_a-1}^{1}\sim\mathcal{N}(0,I)$,
the update is applied from $\tau=1$ to $\tau=0$
to obtain the predicted action chunk
$a_{t:t+H_a-1}^{0}$.
In this work, we use the continuous flow-matching
policy directly, conditioned on the agricultural task
instruction and observations, without intermediate
subtask generation or FAST-token prediction.

\section{Agricultural Environments and Data Collection}
\label{sec: env}

This section describes the simulated and physical agricultural
environments used to implement and evaluate the reaching-and-
grasping task introduced in Section~\ref{subsec:task_formulation}.
We present the robotic platforms, sensor configurations, and
demonstration collection procedures. The simulation environment
supports scalable data collection and controlled evaluation,
while the physical setup enables further policy adaptation
and evaluation under real-world conditions. We also describe
the teleoperation and low-level control framework used for
demonstration collection and policy execution

\subsection{Simulation  Environment}
\label{sec:agri_env}

\begin{figure}[tp]
  \centerline{\includegraphics[width=0.45\textwidth]{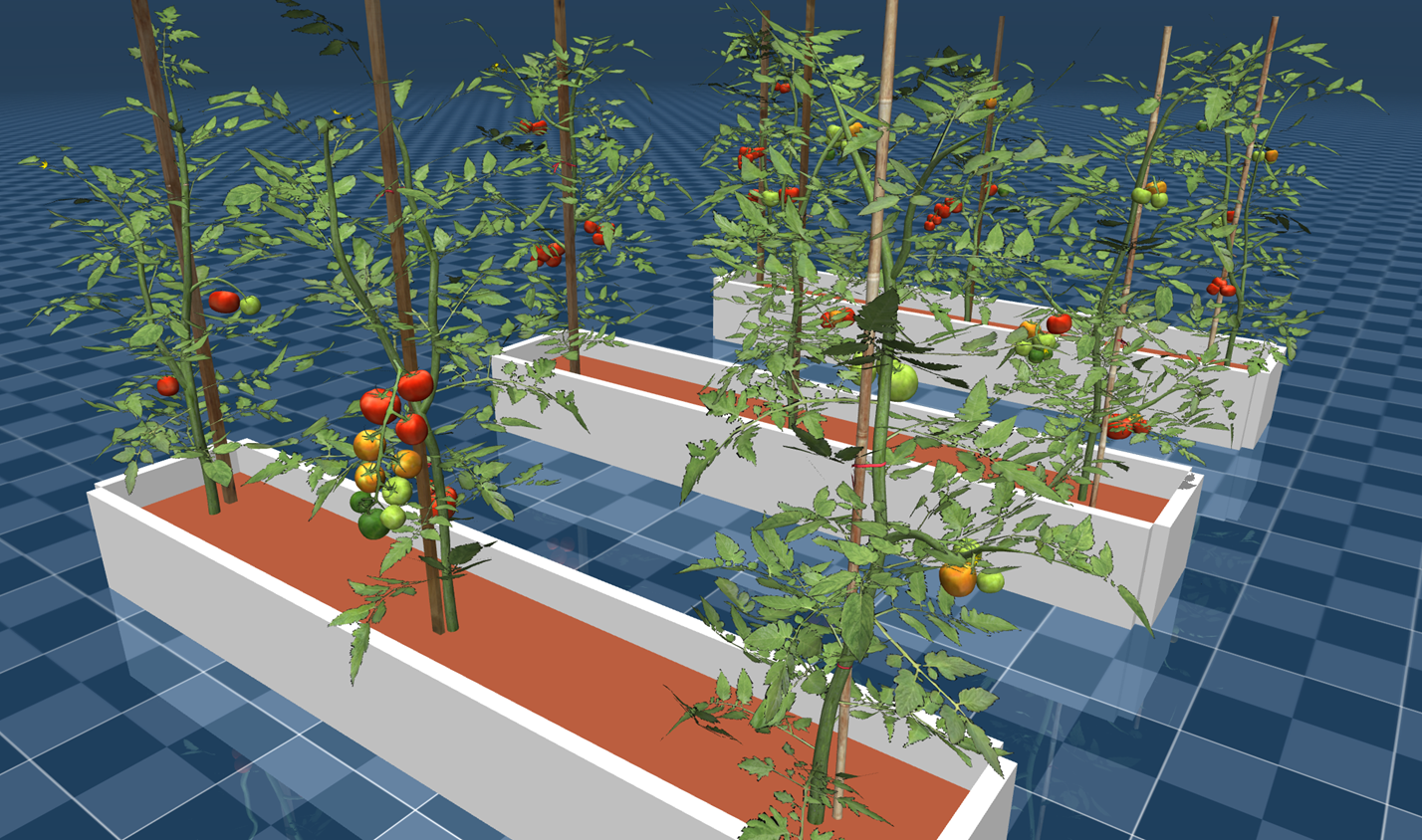}}
  \caption{Nine different tomato plants are used to increase the variability and realism of the agricultural environment.}
  \label{fig:plants}
\end{figure}

To investigate the adaptation of generalist robot policies to
agricultural manipulation, we develop a MuJoCo-based simulation
environment for scalable demonstration collection and controlled
policy evaluation. The environment captures key challenges of
agricultural manipulation, including variations in plant geometry,
clutter from leaves and branches, visual occlusions, and partially
hidden or difficult-to-reach fruits. Nine tomato plant models
with different branch structures, leaf arrangements, fruit
locations, and color variations are incorporated to provide
visual and geometric diversity. As shown in Fig.~\ref{fig:plants},
the plants are arranged in three rows to resemble a greenhouse-like
setting.

The tomato plant models were obtained from existing 3D assets and processed in Blender to make them suitable for MuJoCo simulation. Due to the complex, non-convex geometry of the branches and stems, each plant mesh was decomposed into multiple convex components to obtain collision representations that more closely match the visual geometry. This was necessary because using a single convex hull would produce inaccurate collision boundaries around the irregular plant structures within the Mujoco. Additional processing was required for the leaves, which were originally represented as rectangular planes with textures defining their shapes. These planes were reshaped to approximate the actual leaf contours, avoiding unwanted background surfaces during MuJoCo rendering and improving the correspondence between the visible leaves and their geometry. 
These modifications enabled the integration of visually detailed plant models with more representative collision geometries to support robotic manipulation in cluttered agricultural environments within MuJoCo.


\begin{figure}[tp]
  \centerline{\includegraphics[width=0.5\textwidth]{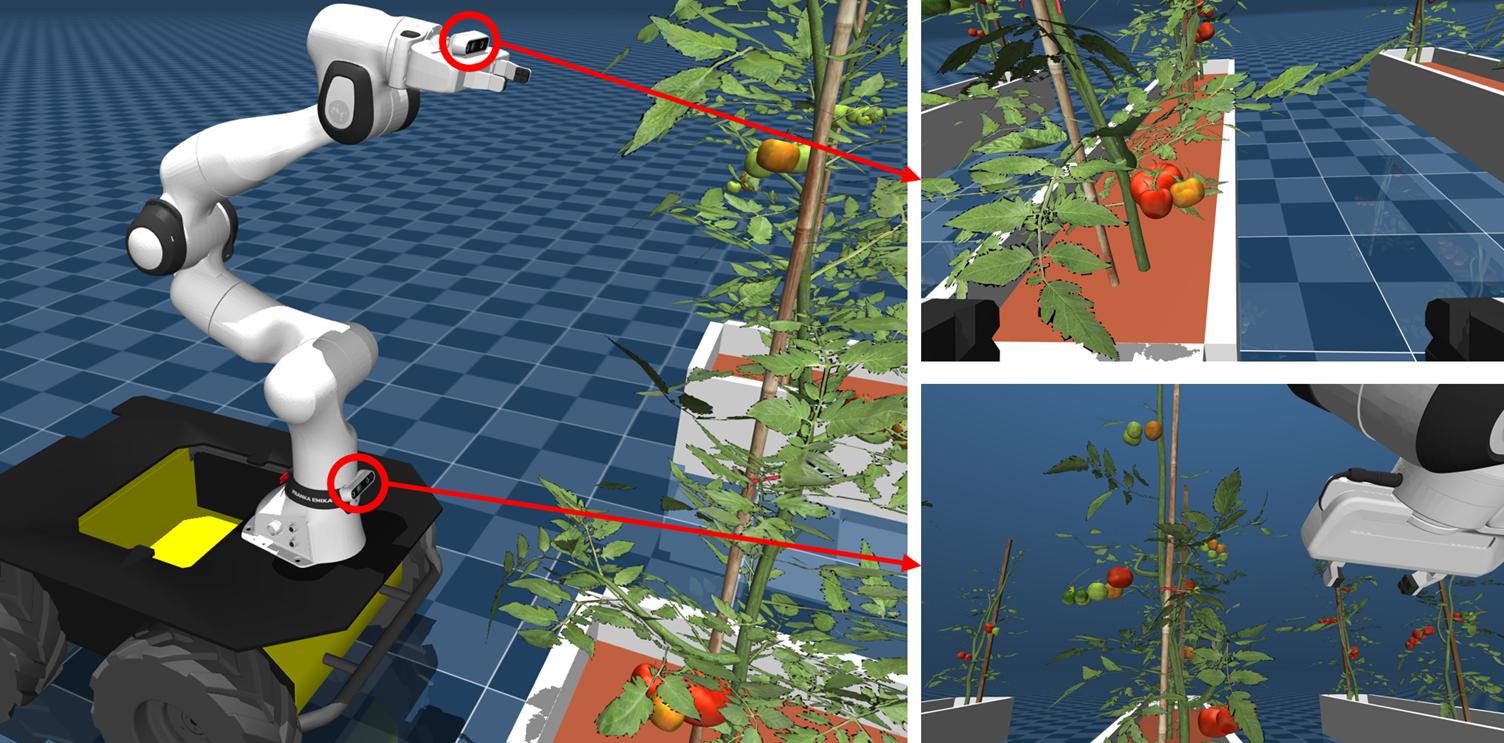}}
  \caption{Simulated mobile manipulator: a 7-DoF FR3 arm mounted on a Husky
  A200 base, with the placement and views of the primary and wrist cameras.}
  \label{fig:robot_sensors}
\end{figure}

As shown in Fig. \ref{fig:robot_sensors}, the simulated robotic platform consists of a Franka Emika Research 3 (FR3) manipulator with seven degrees of freedom and a standard gripper, mounted on a Clearpath Husky A200 mobile base. Although the platform is mobile, its motion is not controlled by the learned policy; the policy generates commands only for the manipulator arm and gripper. The mobile base allows the manipulator to be repositioned near different plants within the agricultural environment. The robot is equipped with two RGB cameras: a wrist-mounted camera positioned near the end-effector and a primary third-person camera mounted near the robot base. The wrist camera provides a close-range view of the end-effector and target region, while the primary camera provides a broader view of the robot and surrounding plant structure. Together, these cameras provide complementary visual information for the policy.

By varying plant geometry, target fruit location, and robot
configuration across episodes, the simulator provides a
controlled yet diverse environment for collecting task-specific
demonstrations and evaluating the fine-tuned Octo and
$\pi_{0.5}$ policies under different agricultural conditions.

\subsection{Real-Robot Experimental Setup}
\label{sec:real_setup}
\begin{figure}[tp]
  \centering
  \includegraphics[width=0.5\textwidth]{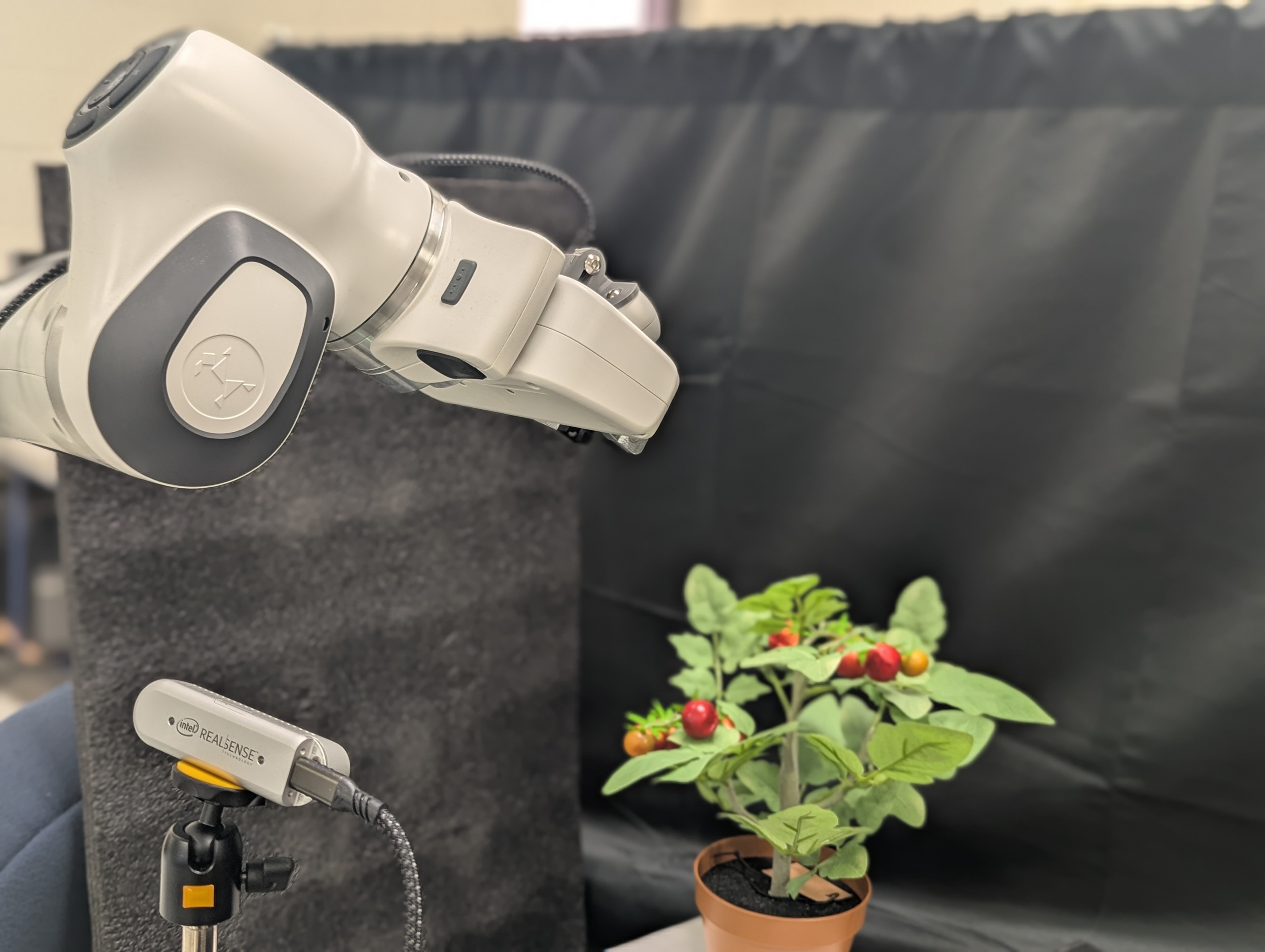}
  \caption{Real-robot experimental setup used for demonstration collection
  and policy evaluation. The FR3 manipulator operates in front of an
  artificial tomato plant using a primary third-person camera and a
  wrist-mounted camera.}
  \label{fig:real_setup}
\end{figure}

To complement the simulation experiments, we develop a
physical agricultural manipulation setup consisting of a
Franka Research 3 (FR3) manipulator equipped with a standard
Franka hand and an artificial tomato plant, as shown in
Fig. \ref{fig:real_setup}. The physical setup maintains the
same reaching-and-grasping task and two-camera observation
configuration used in simulation, while introducing sensing
and execution uncertainties associated with real-world
robot operation.
As in simulation, the physical setup uses a primary third-person RGB camera and a wrist-mounted RGB camera. Unlike the simulated platform, however, the physical FR3 is not mounted on a mobile base. Since the mobile base is not controlled by the VLA policies in simulation, its absence in the physical setup does not change the learned manipulator action space. The primary camera is
positioned independently to provide a broader view of the
workspace and target fruit, while the wrist camera provides a
close-range view of the end-effector and target region. Although
camera placements differ between simulation and the physical
setup, both cameras still provide complementary global and local
visual observations to the policies.

The physical setup is used both to collect real-robot
demonstrations and to evaluate the simulation-adapted
policies after further fine-tuning on these demonstrations. Compared with simulation, the physical system introduces additional sources of uncertainty, including differences in camera viewpoints and image appearance, communication and actuation delays, and variations in gripper execution. These factors allow us to evaluate the adapted
policies under physical operating conditions while maintaining
the same agricultural reaching-and-grasping task.

\subsection{Data Collection}
\label{sec:data_collection}

Two task-specific demonstration datasets are collected for policy adaptation: one in simulation and one on the physical robot. The simulation dataset consists of 200 demonstrations obtained in the MuJoCo agricultural environment, while the real-robot dataset comprises 50 demonstrations obtained using the physical FR3 platform. In both settings, the task is specified by the language instruction ``Reach and grasp the red tomato".

For each demonstration, a human operator controls the robot using a joystick. Demonstrations are recorded at a frequency of $10\,\mathrm{Hz}$. At each time step $t$, the operator specifies a seven-dimensional action
$a = \begin{bmatrix} \delta p^\top & \delta r^\top & g \end{bmatrix}^{\top}$
where $\delta p\in\mathbb{R}^3$ and $\delta r\in\mathbb{R}^3$ denote the translational and rotational end-effector increments, respectively, and $g$ denotes the gripper command. Each action is normalized before being stored in the dataset.


At each time step, synchronized observations from the primary
third-person and wrist-mounted cameras are recorded together
with the robot's proprioceptive state. The proprioceptive
state used for policy fine-tuning comprises seven joint
positions $q=[q_1,\ldots,q_7]^\top\in\mathbb{R}^7$, the
three-dimensional end-effector position, a six-dimensional
representation of the end-effector orientation, and the
gripper state, resulting in a 17-dimensional state vector.
The language instruction and corresponding action are also
recorded at each time step.

For the simulation dataset, demonstrations are collected
across different initial robot configurations, target fruit
locations, and plant geometries. Of the nine simulated tomato
plant models, six are used for demonstration collection,
while the remaining three are reserved for evaluating
generalization to unseen plant structures. The robot platform
is also repositioned between demonstrations to vary the
primary- and wrist-camera viewpoints and introduce different
target configurations and occlusion patterns.

For the real-robot dataset, the same task definition, action
representation, and observation modalities are retained.
The primary-camera viewpoint differs from that used in
simulation, as described in Section~\ref{sec:real_setup}.
Maintaining the same task and action representation across the two domains allows the simulation-adapted policies to be subsequently fine-tuned using the real-robot demonstrations.

\subsection{Low-Level Controller}
\label{sec:controller}

\begin{figure}[tp]
  \centerline{\includegraphics[width=0.5\textwidth]{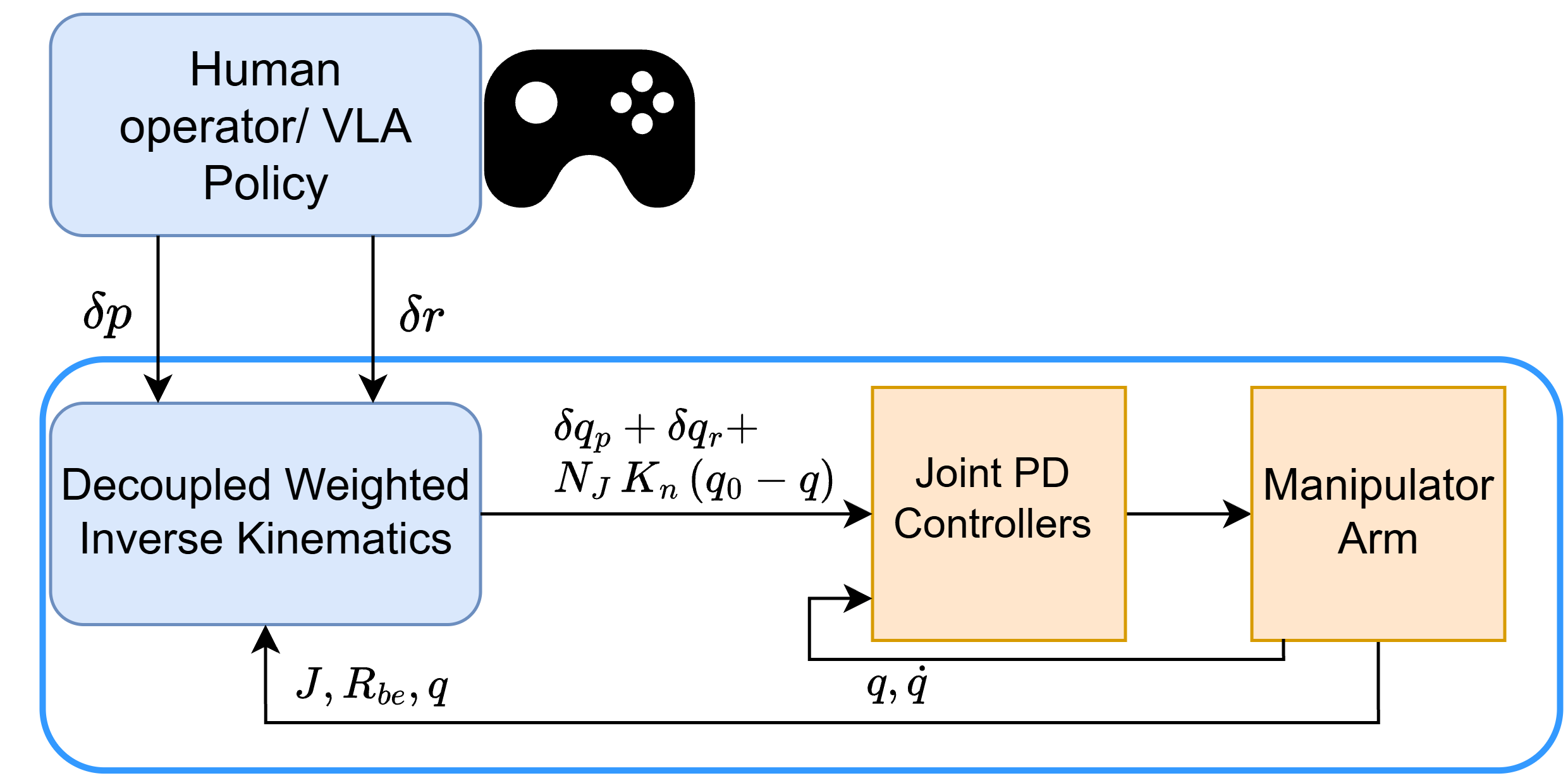}}
  \caption{Low-level controller structure.}
  \label{fig:controller}
\end{figure}

Both the operator's joystick commands and the actions
predicted by the learned policies are executed through the
same low-level control framework (Fig.~\ref{fig:controller}).
Each action specifies an incremental end-effector translation
$\delta p$, expressed in the robot base frame, and an
incremental rotation $\delta r$, expressed in the current
end-effector frame. These commands are converted into
joint-space increments using a damped Jacobian-based
differential inverse-kinematics controller.
For a redundant manipulator such as the FR3, mapping a six-dimensional
end-effector command through the full Jacobian can produce
unintuitive joint motions; for example, a rotation about the
end-effector $z$-axis may involve shoulder and elbow motion
even when the wrist joints could achieve the desired rotation.
To make teleoperation more intuitive for operators and reduce unnecessary arm motion near plant structures, the controller prioritizes translation while biasing rotational motion toward the wrist joints.
The inverse-kinematics controller
determines how the end-effector motion commands are executed
by the manipulator,  while the gripper is controlled independently by mapping the gripper action to an opening or closing command.

Let $J_p, J_r \in \mathbb{R}^{3\times 7}$ denote the translational and
rotational Jacobians expressed in the robot base frame. The translation
command $\delta p$ is already expressed in this frame, while the
rotation command is transformed into the base frame as
$\delta r^{b} = R_{be}\,\delta r$, where $R_{be}$ denotes the
current end-effector orientation relative to the robot base.
The joint increment is computed using a damped pseudo inverse,
\begin{equation}
\delta q_p = J_p^\dagger\,\delta p,
\qquad
J_p^\dagger = J_p^\top\left(J_pJ_p^\top+\alpha I\right)^{-1},
\end{equation}
where $\alpha>0$ is a damping coefficient. The rotational command
is then solved in the null space of the translational task,
$N_{J_p} = I - J_p^\dagger J_p$, to minimize interference with
the commanded translation:
\begin{equation}
\delta q_r = N_{J_p}\,\tilde{J}_r^{\dagger}
\left(\delta r^{b} - J_r\,\delta q_p\right),
\qquad
\tilde{J}_r = J_r N_{J_p}.
\end{equation}
Here, the term $J_r\,\delta q_p$ accounts for the rotation
induced by the translational joint increment, and
$\tilde{J}_r^{\dagger}$ denotes a weighted damped pseudoinverse,
\begin{equation}
\tilde{J}_r^{\dagger}
= W^{-1}\tilde{J}_r^\top
\left(\tilde{J}_r W^{-1}\tilde{J}_r^\top+\alpha I\right)^{-1},
W=\mathrm{diag}(w_1,\dots,w_7),
\end{equation}
with larger weights assigned to the shoulder and elbow joints
than to the wrist joints, encouraging rotational motion to
be produced primarily by the wrist. Finally, the total joint
increment is
\begin{equation}
\delta q = \delta q_p + \delta q_r + N_J\,K_n\,(q_0 - q),
\end{equation}
where $N_J$ is the null-space projection matrix associated
with the full Jacobian, $q_0$ is a nominal joint configuration, and $K_n$ is a diagonal gain matrix. This null-space term exploits the redundancy of the FR3 to keep the arm near a nominal joint configuration while minimizing interference with the commanded end-effector motion. The resulting joint targets $q + \delta q$ are bounded by the joint limits and tracked by joint-level PD controllers.

\section{Fine-Tuning and Evaluation}
\label{section:finetunining_evaluation}

This section describes the fine-tuning and evaluation of Octo and \(\pi_{0.5}\) for the agricultural reaching-and-grasping task. The experimental procedure consists of two stages. First, both policies are fine-tuned using the simulation dataset and evaluated in the MuJoCo agricultural environment. The resulting simulation-adapted policies are then further fine-tuned using the smaller real-robot dataset and evaluated on the physical FR3 platform.
In both simulation and real-robot experiments, we consider two observation settings: one with proprioceptive information and one without proprioceptive input. This allows us to compare the two policy architectures and investigate the effect of proprioceptive information on their performance across simulated and physical environments.

\subsection{Fine-Tuning}
\label{subsec:finetuning}

Octo and $\pi_{0.5}$ are fine-tuned using the same
demonstration datasets within each domain. The simulation
fine-tuning stage uses the 200 demonstrations described in
Section~\ref{sec:data_collection}, while the real-robot
fine-tuning stage uses the 50 demonstrations collected on
the physical FR3 platform. For each model, separate variants
with and without informative proprioceptive input are trained
using the same underlying datasets.

Although both models use the same demonstration data and
are evaluated under the same task conditions, their fine-tuning
procedures follow their respective architectures and
implementations. Consequently, their observation histories,
action-prediction horizons, trainable components, and
optimization schedules are not necessarily identical. All
models are fine-tuned using two NVIDIA H100 GPUs with a
global batch size of $32$. The experiments therefore provide
a practical comparison of the two pretrained policy
architectures rather than a parameter-matched architectural
ablation.

\paragraph{Octo adaptation}

We initialize Octo from the pretrained Octo-Base checkpoint.
The policy uses an observation history of $H_o=2$ time
steps and receives RGB observations from both the primary
and wrist-mounted cameras. In the original Octo configuration,
the primary and wrist-camera observations are processed at
$256\times256$ and $128\times128$ pixels, respectively.
In our experiments, we increase the wrist-camera resolution
to $256\times256$ to use the same input resolution for both
camera streams and improve comparability with the visual
inputs used by $\pi_{0.5}$. 
Because the FR3 action representation differs from those
used during Octo pretraining, the pretrained action head is
replaced with a newly initialized diffusion action head
configured for the seven-dimensional FR3 action space and
an action prediction horizon of $H_a=4$.

For the no-proprioception variant, the underlying demonstration
dataset remains unchanged, but the low-dimensional observation
tokenizer and its corresponding projection are removed from
the Octo input structure. Consequently, no proprioceptive
tokens are provided to the Transformer, and the policy
conditions only on the two visual streams and the language
instruction.

The pretrained T5 language encoder remains frozen during
fine-tuning, while the visual encoder, Octo backbone
Transformer, and newly initialized diffusion action head
are trainable. No image augmentation is applied. The model
is optimized using AdamW, with the learning rate linearly
increased from zero to $1\times10^{-5}$ during the first
100 optimization steps and then held constant. 
The models are first fine-tuned on the simulation dataset for $100{,}000$
optimization steps, after which the resulting simulation-adapted
checkpoints are further fine-tuned on the real-robot dataset
for $20{,}000$ steps.

\paragraph{$\pi_{0.5}$ adaptation}

We initialize $\pi_{0.5}$ from the released pretrained
checkpoint using the LeRobot implementation
\cite{cadenelerobot}. Since our reaching-and-grasping task
uses a single language instruction and does not provide
separate high-level subtask annotations, the policy is
fine-tuned directly for continuous FR3 action prediction.

The policy uses the current observation only, corresponding
to $H_o=1$, and receives RGB observations from the primary
and wrist cameras. Both image streams are stored at
$256\times256$ pixels and resized internally to
$224\times224$ by the model's preprocessing pipeline.
The action-prediction horizon is set to $H_a=10$.

For the no-proprioception variant, the same model structure
is retained, but the processed robot state is replaced with
a constant zero vector before state tokenization. In
$\pi_{0.5}$, the state is discretized and incorporated into
the text prompt together with the language instruction.
Consequently, the resulting state tokens do not convey
information about the current robot state.

The $\pi_{0.5}$ policy is fine-tuned with both the
vision-language backbone and action expert remaining
trainable. No image augmentation is applied. Training uses
AdamW with a peak learning rate of $2.5\times10^{-5}$,
$(\beta_1,\beta_2)=(0.9,0.95)$, weight decay of $0.01$,
and a gradient-clipping norm of $1.0$. A cosine-decay
learning-rate schedule with warmup is used, together with
bfloat16 precision and gradient checkpointing. 
The model are first fine-tuned on the simulation dataset for $6{,}000$
optimization steps, then  this checkpoint is further fine-tuned
on the real-robot dataset for $1{,}000$ steps. At inference,
ten flow-matching integration steps are used to generate
each predicted action chunk.

\subsection{Evaluation}
\label{subsec:evaluation}

\begin{figure*}[t]
    \centering

    \includegraphics[width=0.155\textwidth]{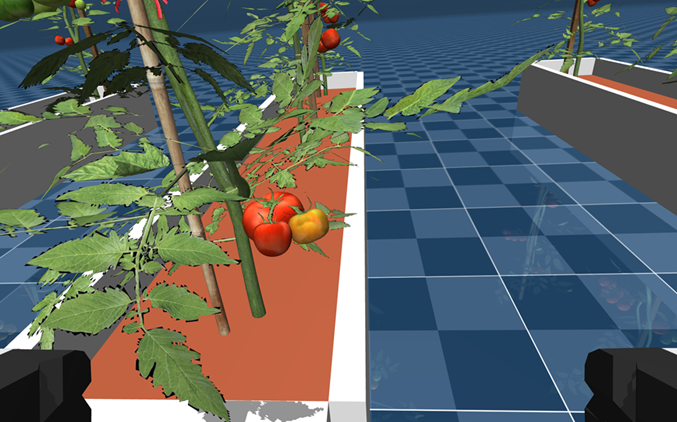}
    \includegraphics[width=0.155\textwidth]{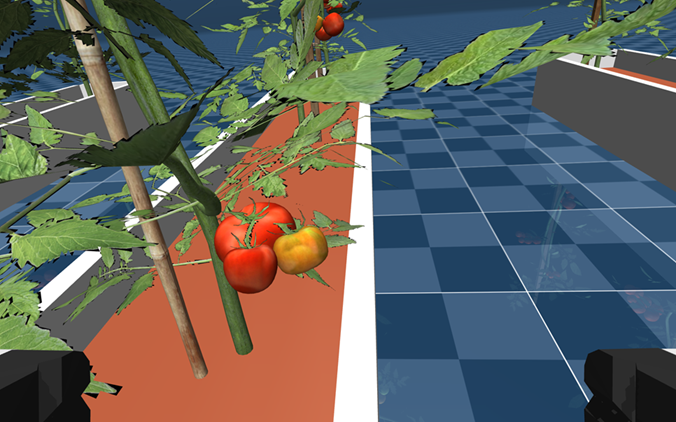}
    \includegraphics[width=0.155\textwidth]{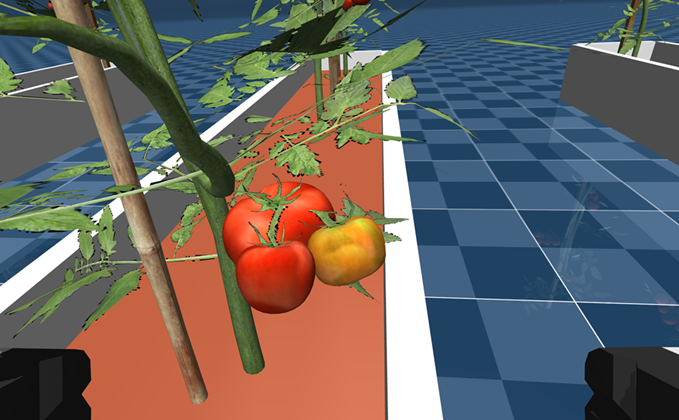}
    \includegraphics[width=0.155\textwidth]{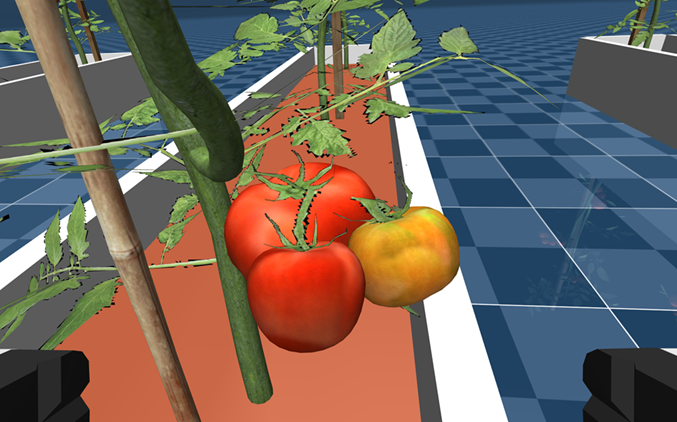}
    \includegraphics[width=0.155\textwidth]{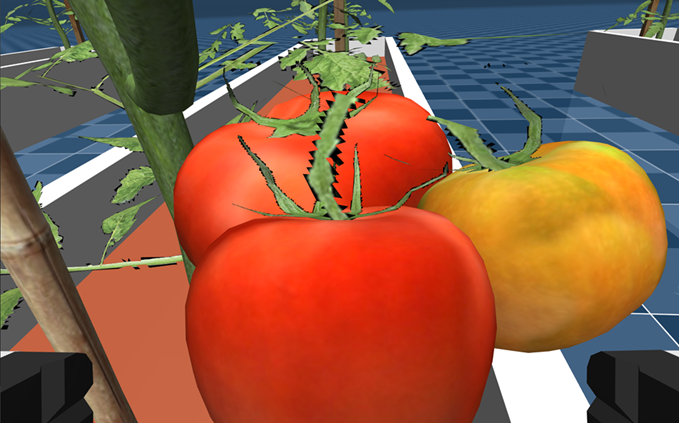}
    \includegraphics[width=0.155\textwidth]{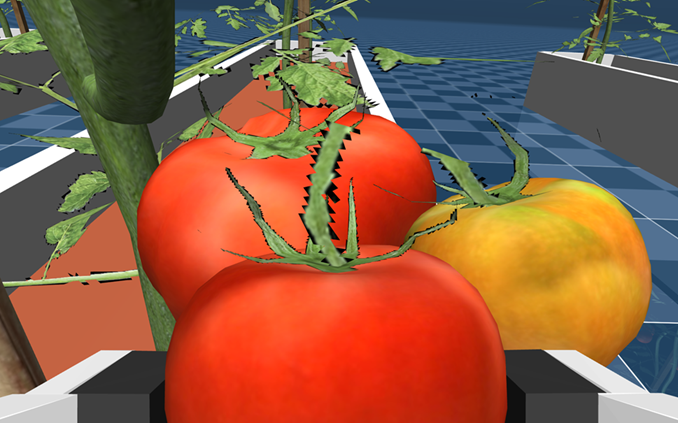}

    \vspace{2mm}

    \includegraphics[width=0.155\textwidth]{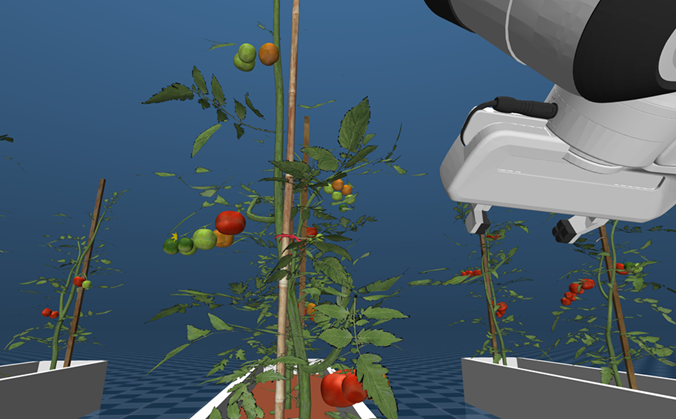}
    \includegraphics[width=0.155\textwidth]{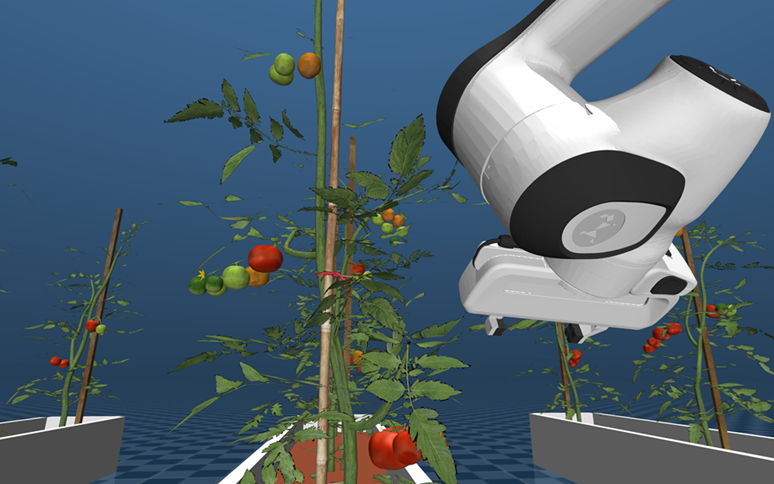}
    \includegraphics[width=0.155\textwidth]{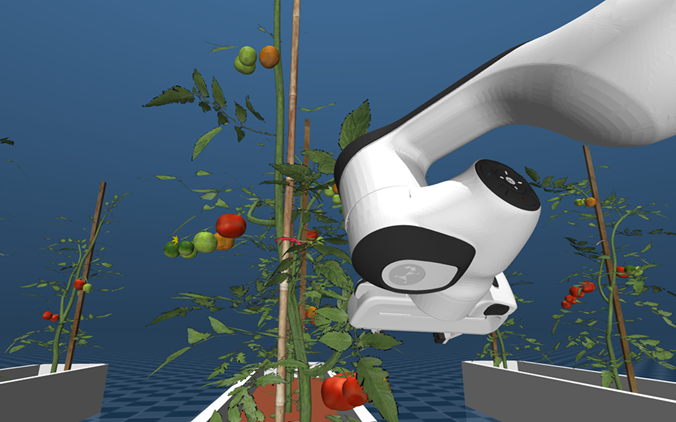}
    \includegraphics[width=0.155\textwidth]{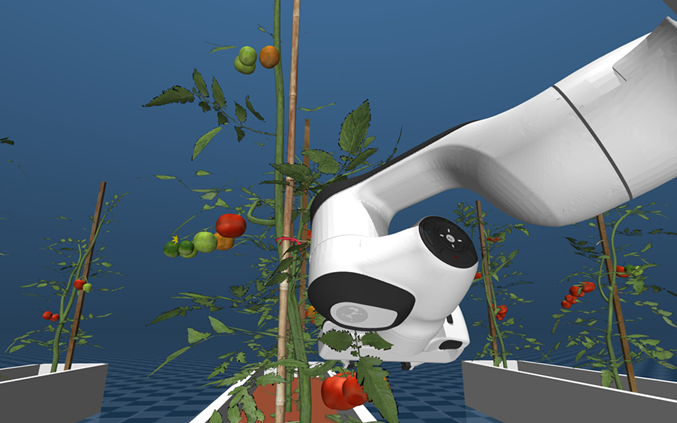}
    \includegraphics[width=0.155\textwidth]{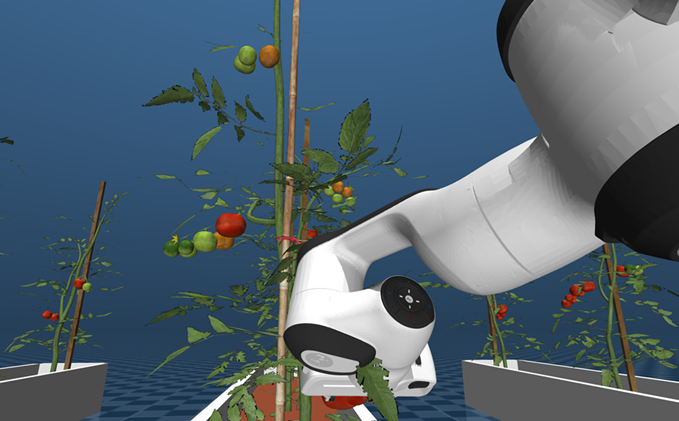}
    \includegraphics[width=0.155\textwidth]{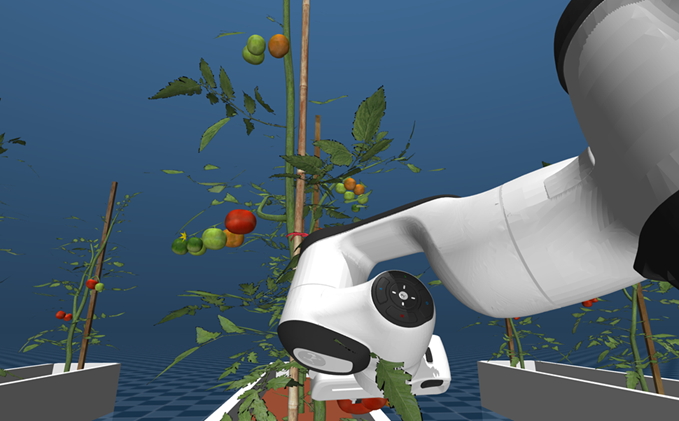}

    \vspace{2mm}

    \includegraphics[width=0.155\textwidth]{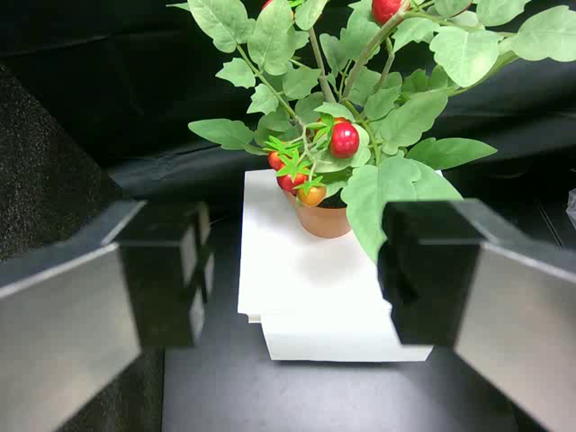}
    \includegraphics[width=0.155\textwidth]{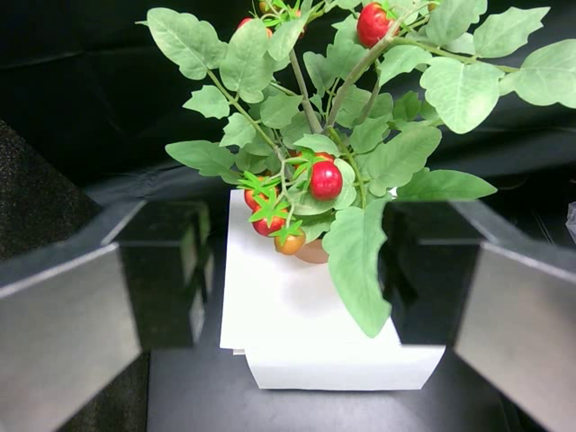}
    \includegraphics[width=0.155\textwidth]{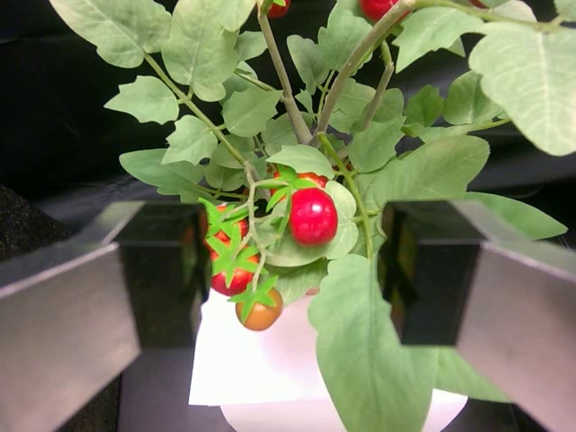}
    \includegraphics[width=0.155\textwidth]{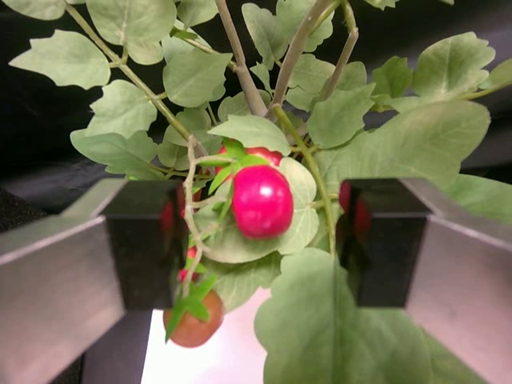}
    \includegraphics[width=0.155\textwidth]{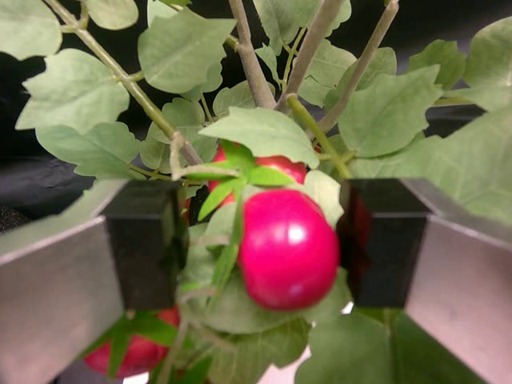}
    \includegraphics[width=0.155\textwidth]{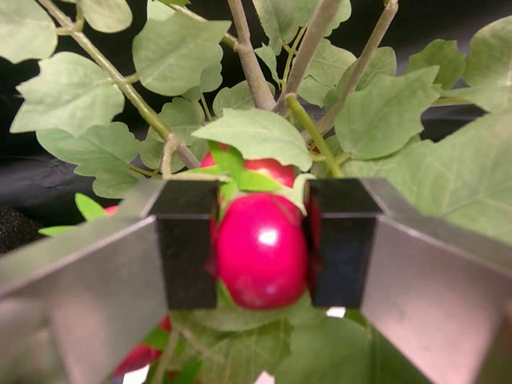}

    \vspace{2mm}

    \includegraphics[width=0.155\textwidth]{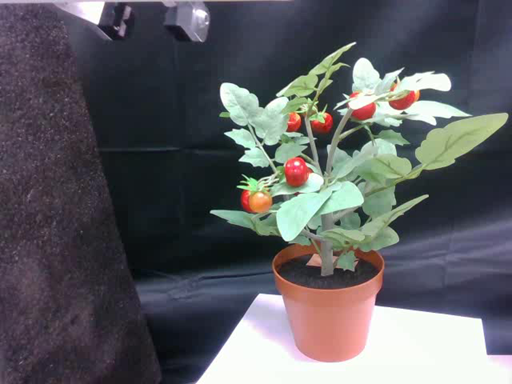}
    \includegraphics[width=0.155\textwidth]{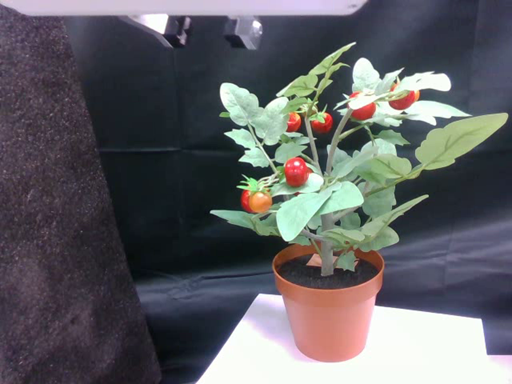}
    \includegraphics[width=0.155\textwidth]{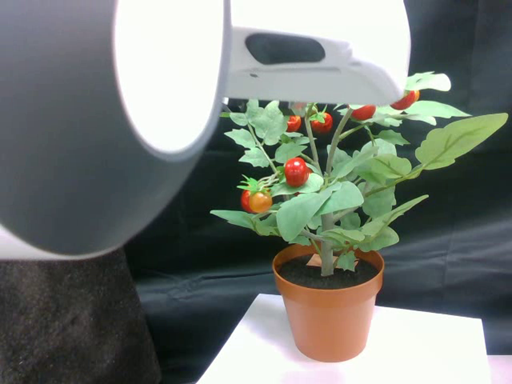}
    \includegraphics[width=0.155\textwidth]{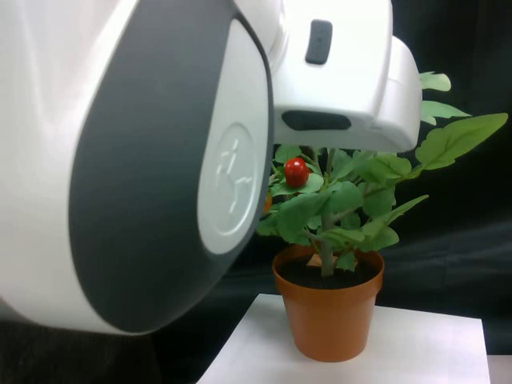}
    \includegraphics[width=0.155\textwidth]{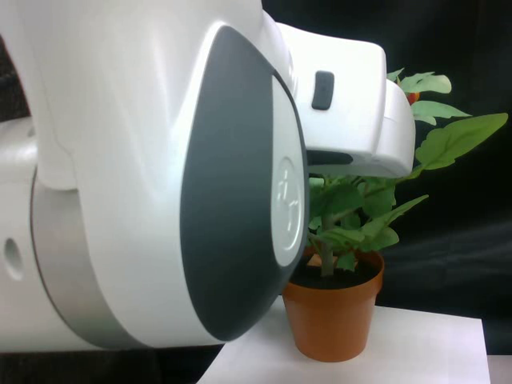}
    \includegraphics[width=0.155\textwidth]{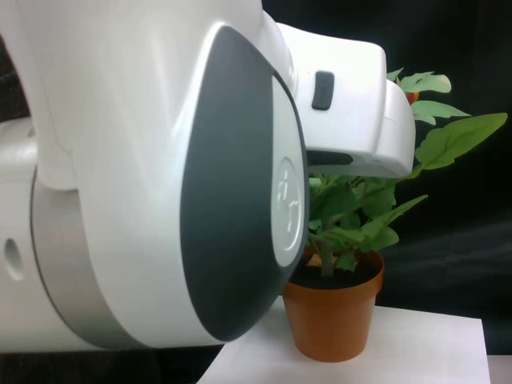}

    \caption{Task executions in simulation and on the physical
robot. The first two rows show the progression of the reaching-and-grasping
task, while the last two rows show the corresponding primary and
wrist camera observations during execution.}
    \label{fig:qualitative_results}
\end{figure*}

We evaluate Octo and $\pi_{0.5}$ in both the simulated
agricultural environment and on the physical FR3 platform.
For each model, the variants with and without informative
proprioceptive input are evaluated separately. 
Within each domain, all policy variants are evaluated
under comparable conditions and using the same task instruction.

During evaluation, both policies employ action-chunk execution,
in which multiple predicted actions are executed consecutively
before the next policy inference. This allows the robot to execute commands at a higher control frequency than the policy inference rate and can promote smoother and more temporally consistent motion. Octo
predicts an action chunk with a horizon of $H_a=4$ and executes
all four actions, whereas $\pi_{0.5}$ predicts a chunk with a
horizon of $H_a=10$ but executes only the first five actions
before generating a new chunk. The same execution strategy
is used in both simulation and real-robot experiments.

An evaluation trial is considered fully successful if the robot
reaches and successfully grasps the target ripe red tomato.
A trial is considered partially successful if the robot reaches
the correct target but fails to complete the grasp, while all
other outcomes are classified as failures. Full success,
partial success, and failure are assigned scores of $100\%$,
$50\%$, and $0\%$, respectively. The reported task performance
score is calculated as the average score across all evaluation
trials. Examples of successful reaching-and-grasping executions
in simulation and on the physical robot, together with
the corresponding primary and wrist camera observations,
are shown in Fig.~\ref{fig:qualitative_results}.


\subsubsection{Simulation Evaluation}

The simulation evaluation considers three scenarios
representing different deviations from the demonstration
distribution: \emph{in-distribution},
\emph{viewpoint/spatial shift}, and
\emph{out-of-distribution}.
The
in-distribution setting uses conditions similar to those in the demonstration
data. The viewpoint/spatial-shift setting uses previously seen plant
instances but perturbs the initial manipulator configuration, producing
different end-effector poses and wrist-camera viewpoints. Finally, the
out-of-distribution setting evaluates generalization to plant structures not
used during fine-tuning. For each policy variant, 25 trials are conducted under each of the three
simulation scenarios. The same evaluation configurations are used across
Octo and $\pi_{0.5}$ and across their proprioceptive and no-proprioception
variants.

\subsubsection{Real-Robot Evaluation}

The real-robot evaluation uses the same reaching-and-grasping task
and compares the same four policy configurations considered in
simulation: Octo with and without proprioception and $\pi_{0.5}$
with and without proprioception. Each policy variant is evaluated
over 10 trials spanning variations in the initial manipulator
configuration, plant placement and orientation, and resulting
camera viewpoints. Although the exact physical configurations
are not matched across policy variants, the trials are selected
to provide approximately comparable task difficulty, particularly with respect to the target tomato and the required reaching and
grasping motion.

\subsection{Results and Discussion}
\label{subsec:results}

\subsubsection{Simulation Results}

\begin{figure}[tp]
  \centerline{\includegraphics[width=0.5\textwidth]{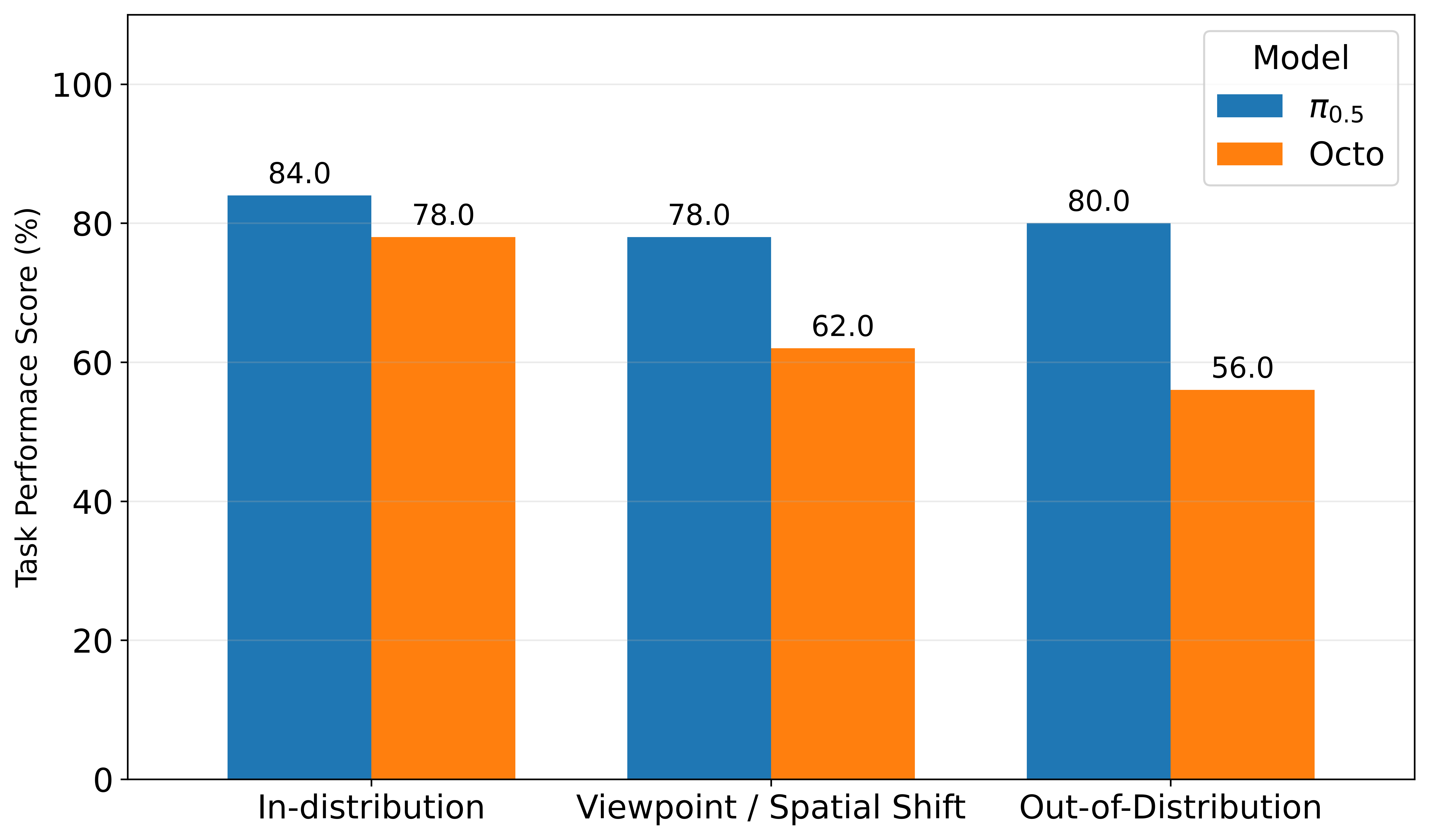}}
  \caption{Octo and $\pi_{0.5}$ performance when proprio is also included in observation based on three scenarios.  }
\label{fig:sim_with_proprio}
\end{figure}

\begin{figure}[tp]
  \centerline{\includegraphics[width=0.5\textwidth]{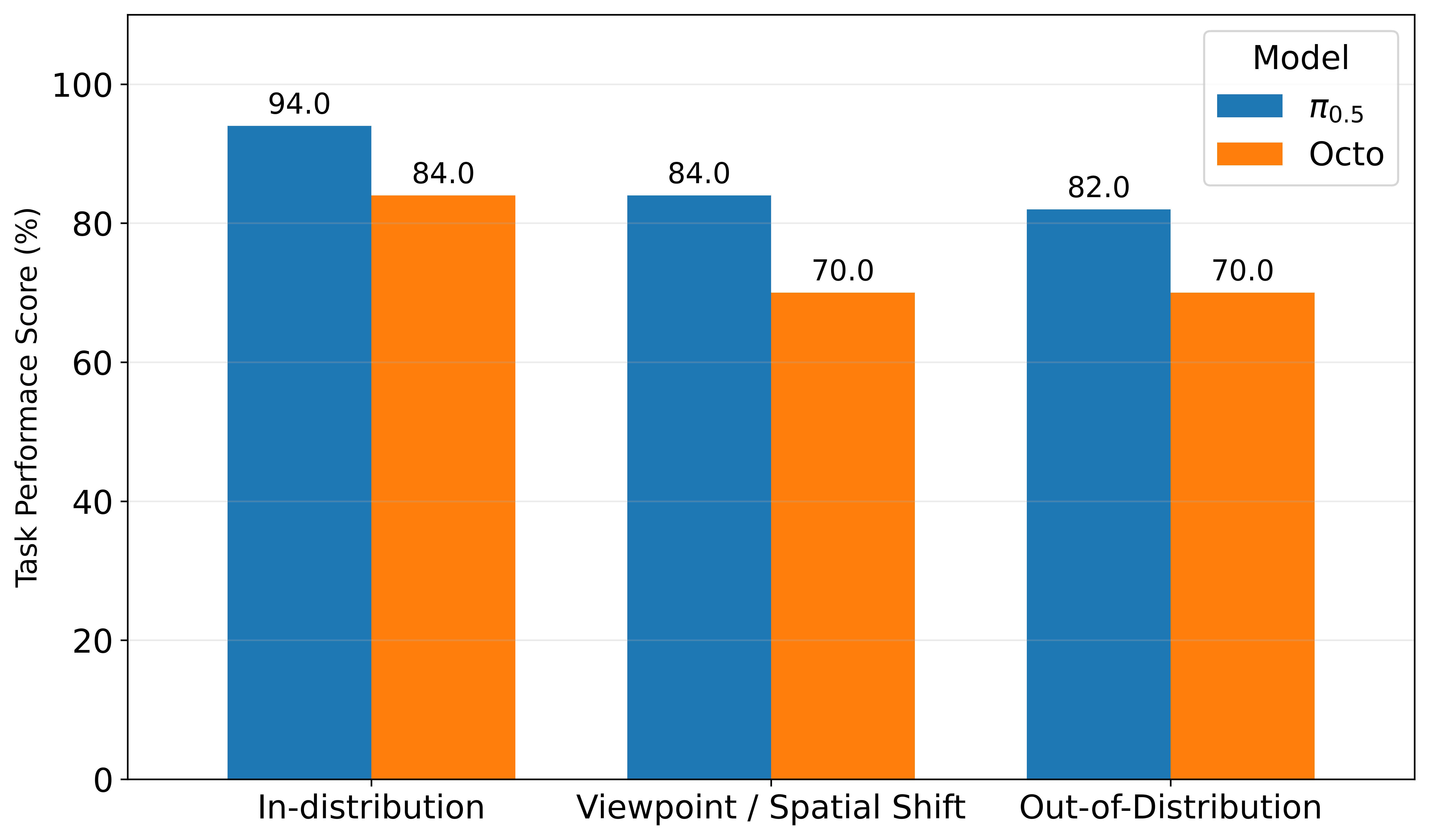}}
  \caption{Octo and $\pi_{0.5}$ performance without proprioception as observation based on three scenarios.}
\label{fig:sim_without_proprio}
\end{figure}

Figs. \ref{fig:sim_with_proprio} and
\ref{fig:sim_without_proprio} compare the task-performance
scores of Octo and $\pi_{0.5}$ across the three simulation
scenarios. With proprioception, $\pi_{0.5}$ achieves scores
of $84\%$, $78\%$, and $80\%$ for the in-distribution,
viewpoint/spatial-shift, and out-of-distribution conditions,
respectively, compared with $78\%$, $62\%$, and $56\%$
for Octo. Without proprioception, the corresponding scores
are $94\%$, $84\%$, and $82\%$ for $\pi_{0.5}$ and
$84\%$, $70\%$, and $70\%$ for Octo.

Across both observation settings, $\pi_{0.5}$ maintains
higher task-performance scores than Octo in all three
simulation scenarios. The difference is particularly
pronounced for the viewpoint/spatial-shift and
out-of-distribution conditions, where Octo exhibits a
larger reduction relative to its in-distribution
performance. 
These results highlight the importance of handling variations in plant structure and robot viewpoint in agricultural manipulation, where fruit locations and branch and leaf arrangements can differ across individual plants.
The higher performance of
$\pi_{0.5}$, however, is associated with
higher computational demands, as $\pi_{0.5}$ employs a larger pretrained vision-language backbone
and generally requires more computation
at inference than Octo. Therefore, the two models represent
potential trade-offs between policy performance and
computational efficiency. While the
additional capacity of $\pi_{0.5}$ may be particularly
useful when extending the system to multiple tasks or
more diverse language instructions, this capability is
not directly evaluated in the present single-task
experiments.

\begin{figure}[tp]
  \centerline{\includegraphics[width=0.5\textwidth]{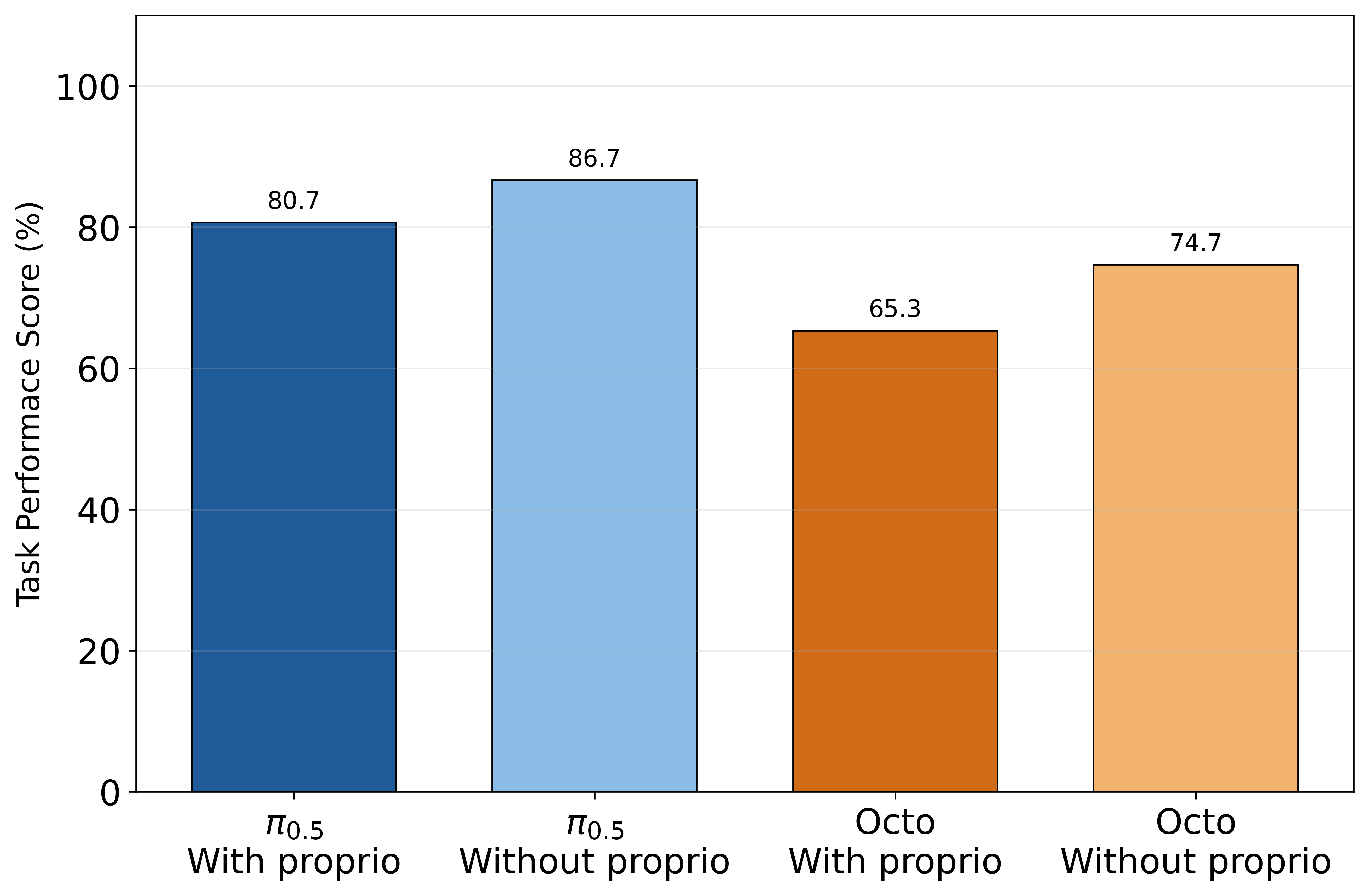}}
  \caption{Average performance of Octo and $\pi_{0.5}$ with and without proprioception as observation.}
\label{fig:sim_average}
\end{figure}

A second consistent observation is that removing
proprioceptive information improves the performance of
both models. As summarized in Fig. \ref{fig:sim_average},
the average task-performance score of $\pi_{0.5}$
increases from $80.7\%$ with proprioception to $86.7\%$
without proprioception, while that of Octo increases
from $65.3\%$ to $74.7\%$.

These results suggest that the additional low-dimensional state information
does not provide a performance advantage for the reaching-and-grasping task
under the training conditions considered here. One possible explanation is
that the two camera images already capture substantial information about the
relative configuration of the end-effector, target fruit, and surrounding
plant structure, reducing the benefit of additional state information.
Moreover, proprioceptive input may allow the policy to exploit correlations
between robot configurations and demonstrated actions rather than relying
primarily on visual relationships to the target. This possibility is related
to the causal confusion problem in imitation learning \cite{de2019causal},
where additional correlated observations can lead to poorer generalization
under distribution shift. While our experiments do not directly establish
this mechanism, the results are consistent with this possibility. For
example, Octo shows its largest improvement from removing proprioception
under the out-of-distribution condition, increasing from $56\%$ to $70\%$.
Importantly, this result should not be interpreted as indicating that
proprioception is generally detrimental to VLA policies. Its usefulness
depends on the task, state representation, model architecture, training
distribution, and available adaptation data. Rather, in the present
experiments, both models achieve higher task-performance scores without
proprioceptive input.

\subsubsection{Real-Robot Results}

\begin{figure}[tp]
  \centerline{\includegraphics[width=0.5\textwidth]{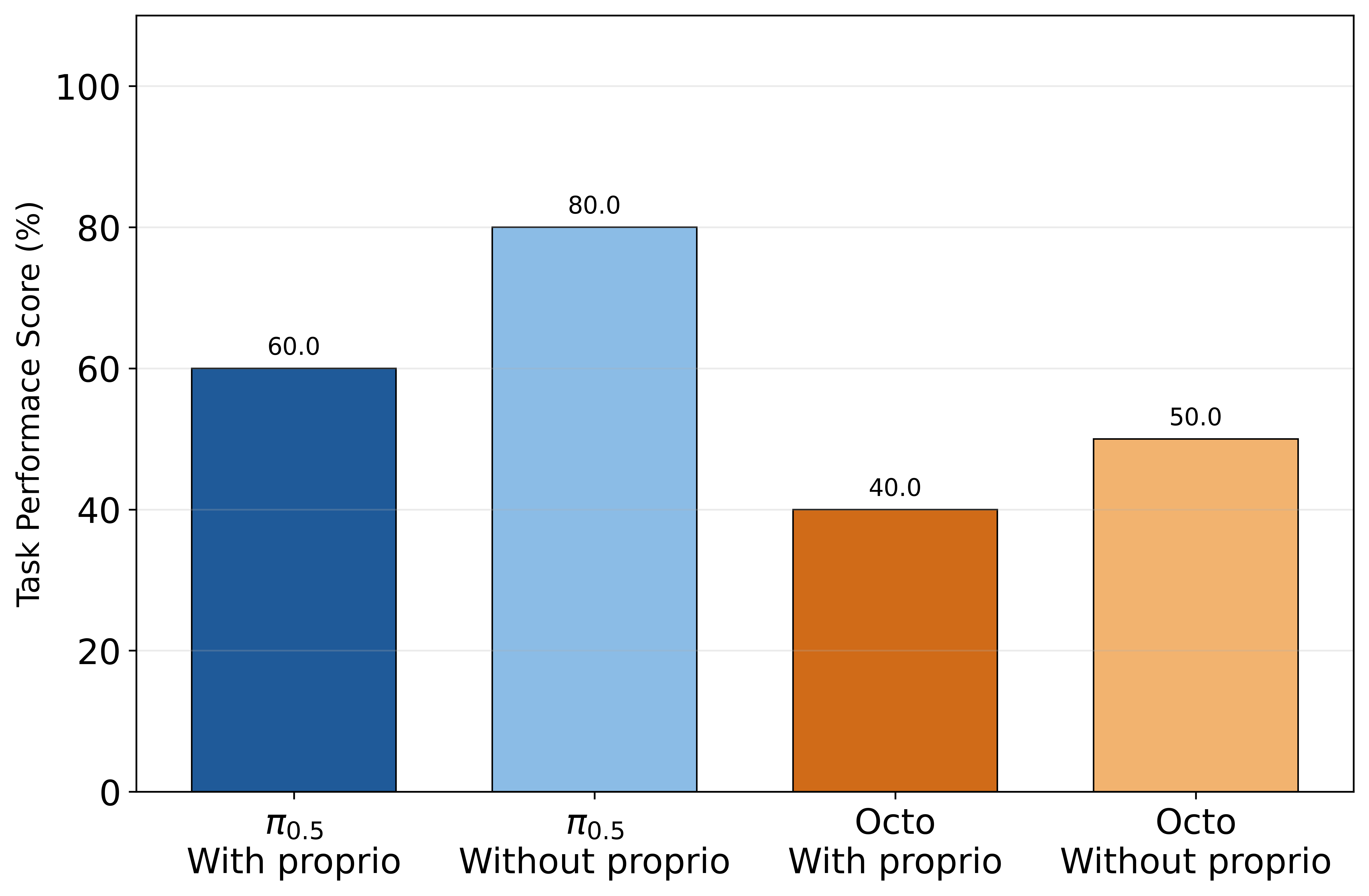}}
  \caption{Performace of Octo and $\pi_{0.5}$ with and without proprioception as observation for real task.}
  \label{fig:real_comparison}
\end{figure}

The general trends observed in simulation are also reflected
in the real-robot experiments, as shown in
Fig.~\ref{fig:real_comparison}. $\pi_{0.5}$ achieves
task-performance scores of $60\%$ with proprioception
and $80\%$ without proprioception. For Octo, the
corresponding scores are $40\%$ and $50\%$, respectively.

Although the task-performance scores are generally lower
on the physical robot than in simulation, the same overall
trends are observed. $\pi_{0.5}$ achieves higher scores
than Octo, while the variant without proprioception
achieves a higher score than the corresponding variant
with proprioception for each model. 
These results are consistent with the relative performance
differences observed in simulation, although the limited
number of physical trials and differences in evaluation
configurations should be considered when interpreting
the comparisons. The physical system also introduces
additional challenges that are not fully represented
in simulation. 

Several factors may contribute to the lower real-robot
performance. In the physical setup, branches and fruits
are compliant and can move when contacted by the
manipulator, causing the target pose to change during
the final approach and grasp. The physical gripper also
exhibits slower and more variable actuation than its
simulated counterpart, making grasp timing more critical.
In addition, sensing, communication, and control delays
affect the timing between observation, policy inference,
and action execution. These effects can have a greater
impact on the physical robot than in simulation, where
sensing and actuation are more deterministic. Consequently,
small positioning or timing errors near the target can
lead to unsuccessful grasps even when the policy
successfully reaches the correct fruit.


Despite these challenges, the experiments highlight a
potential practical advantage of pretrained generalist
robot policies for agricultural manipulation. In
conventional modular pipelines, adapting a system to
new plant geometries or camera configurations may
require adjustments of perception, planning,
and control components, such as updating fruit-detection
models, camera calibration, or geometric representations
used for motion planning
\cite{miao2023efficient,gursoy2023towards}.
In contrast, the policies in this work were first adapted
using simulated demonstrations and subsequently fine-tuned
with only 50 real-robot demonstrations, despite differences
in camera placement, scene appearance, and plant
characteristics between the two environments.
Adaptation was thus achieved primarily through additional
human demonstrations rather than extensive redesign and
retuning of individual perception and planning components.
Although real-robot fine-tuning and reliable low-level
control remain necessary, these results suggest that
pretrained generalist robot policies offer a promising
approach to simplifying the adaptation of robotic
manipulation systems to variable agricultural environments.


\section{Conclusion}

This work investigates the potential of pretrained generalist
robot policies for agricultural manipulation, using tomato reaching
and grasping as a representative task. A MuJoCo-based agricultural
environment was developed to support demonstration collection and
controlled evaluation under variations in plant structure, robot
configuration, and viewpoint. The Octo and $\pi_{0.5}$ policies
were adapted using simulated demonstrations, followed by further
fine-tuning with real-robot demonstrations and evaluation on a
physical FR3 manipulator. In addition, both models were evaluated
with and without proprioceptive information to examine the
contribution of low-dimensional robot state information.

The results indicate that pretrained generalist robot policies
can be successfully adapted to this agricultural manipulation
task in both simulation and on the physical robot. Across the
evaluated conditions, $\pi_{0.5}$ achieved higher task-performance
scores than Octo and showed smaller performance degradation under
the tested distribution shifts in simulation. However, this
performance advantage is associated with higher computational
demands, highlighting potential trade-offs between policy
performance and computational efficiency. The experiments also
showed that including proprioceptive information did not improve
performance for either model in the present task. Instead, both
policies achieved higher scores without proprioception, suggesting
that the benefit of additional observation modalities depends
on the task, state representation, and model architecture.

The real-robot experiments further highlight both the potential
and the remaining challenges of applying pretrained robot policies
to agricultural manipulation. Although the same general performance
trends observed in simulation were maintained on the physical
platform, overall performance was lower, potentially due to
factors that are difficult to fully reproduce in simulation,
including motion and compliance of plant branches and fruits,
variations in gripper execution, and sensing, communication,
and control delays. These results suggest that pretrained
generalist robot policies provide a promising foundation for
agricultural robotic manipulation, while further improvements
in physical interaction modeling, real-world adaptation,
and execution robustness are needed for reliable deployment.

Future work will extend the current single-task setting toward multiple
agricultural manipulation tasks and more diverse plant configurations.
This includes increasing the amount and diversity of real-robot data,
improving the modeling of deformable plant interactions, and
investigating the integration of force and tactile feedback into
policy training to enable gentle, damage-free fruit manipulation.
We will also investigate adaptation strategies that can better
leverage large pretrained robot policies while maintaining
efficient and reliable real-time execution.










\bibliographystyle{IEEEtran}
\bibliography{IEEEexample}

\end{document}